\RequirePackage{fix-cm}
\documentclass[twocolumn]{svjour3}           % two column
\smartqed  % flush right qed marks, e.g. at end of proof
\usepackage{graphicx}
\usepackage{amsfonts,amssymb}
\usepackage{algorithm}
\usepackage{algorithmic}
\usepackage{amsmath}
\usepackage[numbers]{natbib}
\usepackage{color}
\newcommand{\INPUT}{\item[\myinput]}
\newcommand{\myinput}{\textbf{Initialization:}}

\newcommand{\MYWHILE}{\item[\mywhile]}
\newcommand{\mywhile}{\textbf{repeat}}

\newcommand{\MYENDWHILE}{\item[\myendwhile]}
\newcommand{\myendwhile}{\textbf{until}}
\usepackage{mathptmx}      % use Times fonts if available on your TeX system
\usepackage{latexsym}
\usepackage{url}
\usepackage{amsmath}
\usepackage{lineno}
\journalname{International Journal of Computer Vision}
\begin{document}
%\pagewiselinenumbers
\title{A Deep Structured Model with Radius-Margin Bound for 3D Human Activity Recognition%\thanks{Grants or other notes
}

%Here maybe we should highlight the radius bound ??


\author{Liang~Lin \and Keze~Wang \and Wangmeng~Zuo \and Meng~Wang \and Jiebo~Luo \and Lei~Zhang
}

\institute{L. Lin and K. Wang are with the Sun Yat-sen Unviersity and The Hong Kong Polytechnic University, China (e-mail: linliang@ieee.org). \\ W. Zuo is with the School of Computer Science and Technology, Harbin Institute of Technology, Harbin 150001, China (e-mail: cswmzuo@gmail.com). \\ M. Wang is with Hefei University of Technology, China. (e-mail: eric.mengwang@gmail.com). \\ J. Luo is with Department of Computer Science, University of Rochester, U. S. (e-mail: jluo@cs.rochester.edu). \\L. Zhang are with Department of Computing, The Hong Kong Polytechnic University. (e-mail: cslzhang@comp.polyu.edu.hk). }


\date{Received: date / Accepted: date}

\maketitle

\begin{abstract}

%
%Recently developed 3D/depth sensors have opened up new opportunities (e.g. personal assistive robotics) with enormous commercial values, which provide more rich information (e.g. extra depth data of scenes and objects) compared with traditional cameras. However, understanding human activity is still challenging even with these enriched data.

%Understanding human activity is very challenging even with the recently developed 3D/depth sensors. There are two main difficulties: i) the complexity of modeling person appearance and motion and ii) the ambiguity in the temporal segmentation of the sub-activities that constitute an activity.

Understanding human activity is very challenging even with the recently developed 3D/depth sensors. To solve this problem, this work investigates a novel deep structured model, which adaptively decomposes an activity instance into temporal parts using the convolutional neural networks (CNNs). Our model advances the traditional deep learning approaches in two aspects. First, { we incorporate latent temporal structure into the deep model, accounting for large temporal variations of diverse human activities. In particular, we utilize the latent variables to decompose the input activity into a number of temporally segmented sub-activities, and accordingly feed them into the parts (i.e. sub-networks) of the deep architecture}. Second, we incorporate a radius-margin bound as a regularization term into our deep model, which effectively improves the generalization performance for classification. For model training, we propose a principled learning algorithm that iteratively (i) discovers the optimal latent variables (i.e. the ways of activity decomposition) for all training instances, (ii) { updates the classifiers} based on the generated features, and (iii) updates the parameters of multi-layer neural networks. In the experiments, our approach is validated on several complex scenarios for human activity recognition and demonstrates superior performances over other state-of-the-art approaches.

\keywords{Human Action and Activity \and RGB-Depth Analysis \and Structured Model \and Deep Learning}
\end{abstract}

\section{Introduction}
\label{intro}

In computer vision, it has received increasing attention in human activity understanding to determine what people are doing given an observed video in different application domains, e.g. intelligent surveillance, robotics, and human - computer interaction. Recently developed 3D/depth sensors have opened up new opportunities with enormous commercial values, which provide more rich information (e.g. extra depth data of scenes and objects) compared with the traditional cameras. Built upon the enriched information, human poses can be estimated more easily. However, modeling complicated human activities still remains challenging, mainly due to the following difficulties.

%This paper focuses on recognizing complex human activities from Grayscale-Depth videos which are captured by a depth camera (e.g., Microsoft Kinect). Despite the additionally provided depth information, there still exist two main difficulties:

\begin{figure}[!htb]
\centering
\includegraphics[width = \columnwidth]{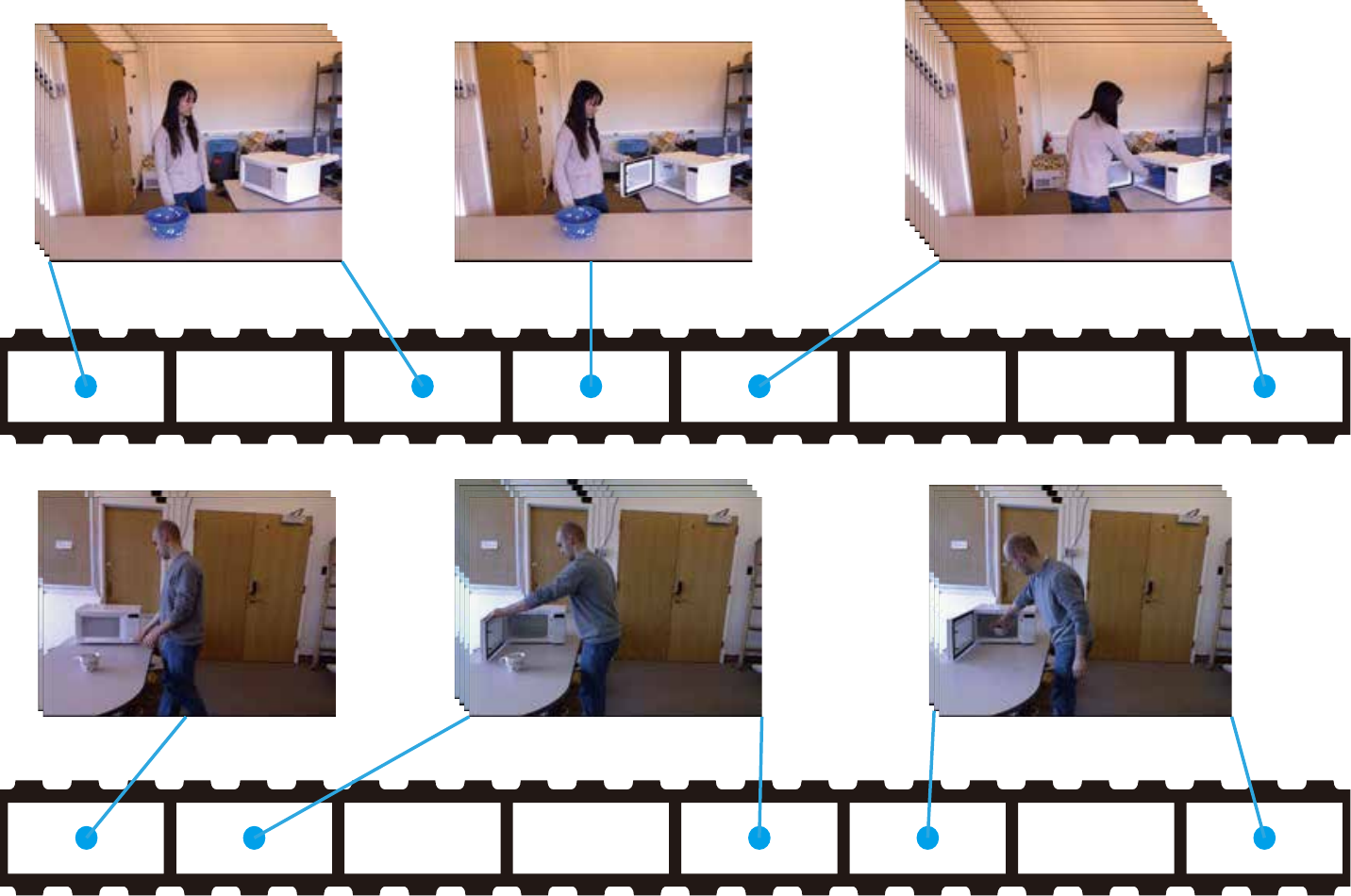}
\caption{Two activities of the same category. We consider one activity as a sequence of actions occurred over time, and the temporal composition of an action may differ for different subjects.}\label{fig:motivation}
\end{figure}

\begin{itemize}
    \item ({\textbf a}) The complexity of representing high-level activities with the rich appearance and motion information from video. The actors may appear in diverse views or poses under different motions, and the surrounding objects and environments can also vary within the same activity category. Moreover, the depth maps provided by the 3D sensors are often unavoidably contaminated \citep{HON4D} due to the { noise} or the self-occlusion of the body parts.

    \item ({\textbf b}) The ambiguity in the temporal segmentation of the sub-activities which constitute an activity. An activity can be considered as a sequence of actions (i.e. sub-activities) occurred over time \citep{CIVU2013survey}. For instance, the activity of ``microwaving food'' can be temporally decomposed into several parts such as picking up food, walking and operating microwave. However, the activity composition may vary for a category of activity instances. Figure~\ref{fig:motivation} shows two activities belonging to the same category, where the temporal lengths of decomposed actions are different for different subjects. It is therefore difficult to capture the temporal variation of activities during the category recognition.
\end{itemize}

Most of previous methods recognize 3D human activities by training discriminative/generative classifiers based on carefully designed features~\citep{HOJ3D,HON4D,DSTIP,WuYingCVPR2012}. These approaches often require sufficient domain knowledge and heavy feature engineering because of the difficulty ({\textbf a}), which could limit their applications. To improve the discriminative performance, some compositional methods~\citep{WangPAMI2011,CIVU2013survey} model complex activities by segmenting the videos into temporal segments of fixed length. But because of the difficulty ({\textbf b}), they may have problems handling complex activities composed of actions of diverse temporal durations, e.g. the examples in Figure~\ref{fig:motivation}.

In this work, { we develop a deep structured human activity model to address the above mentioned challenges}, and demonstrate superior { performance} over other state-of-the-art approaches on the task of recognizing human activities from Grayscale-Depth videos which are captured by a {RGB-D} camera (i.e. Microsoft Kinect). Our model adaptively represents the input activity instance as a sequence of temporally separated sub-activities, and each one is associated with a cubic-like video segment of a flexible length. Our model is inspired by the effectiveness of two widely successful techniques: deep learning \citep{CNN1990,Hinton06,ImagenetNIPS2012,3DCNNPAMI, MDLACM13, PPDDNN13,WangMM2014} and the latent structured models \citep{AOGZhu2006,lsvm-pami,SinisaSPNCVPR2012,AOGICCV2011,DisAOG-PAMI}. One example of the former is Convolutional Neural Networks (CNNs), which was recently applied to generate powerful features for video classification { \citep{3DCNNPAMI, VideoCNNs}}. On the other hand, { the latent structured models (such as Deformable Part-based Model \citep{lsvm-pami}) have been demonstrated as an effective class of models for handling large object variations for recognition and detection. One of the key components in these models is the reconfigurable flexibility of model structure, which often implemented by estimating latent variables during inference.}

\begin{figure*}[!ht]
\centering
\includegraphics[width=7.0in]{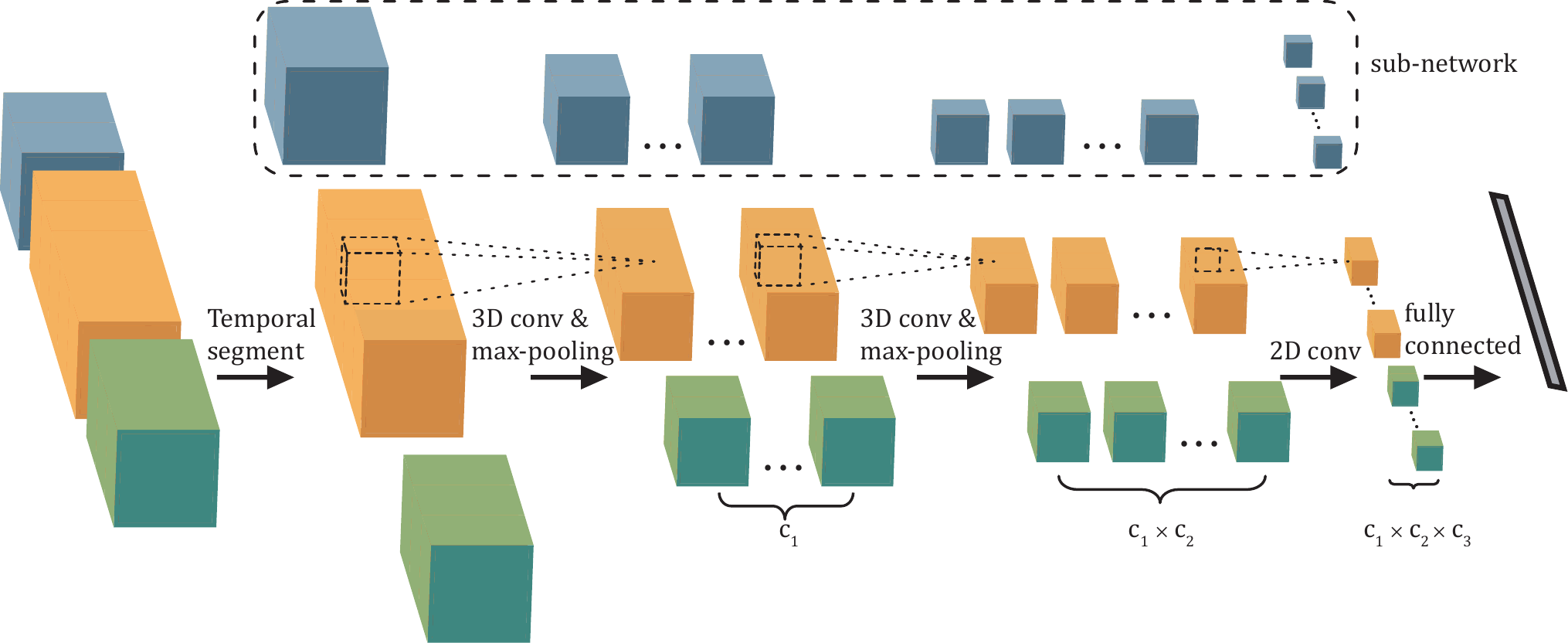}
\caption{ The architecture of spatio-temporal convolutional neural networks.  The neural networks are stacked up by convolutional layers, max-pooling operators and a fully connected layer, where the raw segmented videos are treated as the input. { A sub-network is referred to a vertically-decomposed subpart stacked up by several layers, which extracts features for one segmented video section (i.e. one sub-activity). Moreover, by using the latent variables, our architecture is capable of explicitly handling diverse temporal compositions of complex activities.} }\label{fig:Architecture}
\end{figure*}

We adopt the deep CNN architecture \citep{CNN1990,3DCNNPAMI} to layer-wisely extract features from the input video data, and { the architecture are vertically decomposed into several sub-networks corresponding to the video segments}, as Figure \ref{fig:Architecture} illustrates.  In particular, our model searches for the optimal composition for each activity instance during the recognition, which is the key to handle the temporal variation of human activities. Moreover, we introduce relaxed radius-margin bound into our deep model, which effectively improves the generalization performance for classification. In the following, we briefly overview the main components of our model and summarize the advantages.

%First, our model is capable of performing 3D human activity recognition directly on grayscale-depth data rather than relying on hand-crafted features. Our conneural networks

%{We build the layered neural networks stacked up by convolutional layers, max-pooling operators and a fully connected layer, where the video data are fed to the bottom layer. We firstly apply the 3D convolutional kernel~\citep{3DCNNPAMI} over the bottom to extract features from both spatial and temporal domains, thereby encoding the motion information over adjacent frames. The 2D convolutions are then deployed to extract the higher-level information.}

%
% giving rise to the activity classification. The top layer is kernelized Support Vector Machine (SVM), which can achieve large performance gain than widely used softmax due to the superior regularization effects of the SVM loss function.

% some problems for the model structure, we have two full connection layers

First, the configuration of our deep model can be flexibly adjusted to adapt to different input videos, and the significance of this property has been justified for human action recognition \citep{LiangMM2013,FeifeiCVPR2012, WangMM2014}. In our approach, we make our model adaptively capture temporal structure by using the latent variables. { This motivation finely accords with a batch of existing part-based structured models in visual recognition \citep{DisAOG-PAMI,SwitchDNNs}. More specifically, we utilize the latent variables to explicitly represent the temporal composition of the human activities, i.e. the input video is partitioned into several segments of alterable lengths (each segment indicating a sub-activity). The different temporal compositions actually correspond to the different temporal durations of the separated sub-activities}. And the frames of different video segments are extracted to feed to the corresponding sub-networks.

 { During the inference of activity recognition, we aggregate the responses from sub-networks while searching for the optimal temporal activity segmentation. This inference will inevitably cause extra computation cost just like traditional latent structured models~\citep{DisAOG-PAMI}. It is worth mentioning that we can implement the inference in a parallel manner using GPU (Graphic Processing Unit) programming, in order to counter-balance the extra computational demand.}

Second, we integrate the radius-margin regularization with the deep feature learning, effectively conducting the classification with good generalization performance. Collecting 3D data of human activities is relatively expensive in practice, while the large  { amount} of training data plays a critical role in recent successful deep learning approaches \citep{ImagenetNIPS2012,SwitchDNNs,VideoCNNs}. On the other hand, the max-margin methods (e.g. Support Vector Machines) have shown very impressive generalization power and thus been widely applied for small scale training data. { According to \citep{R-SVM,Radius}, their performance (i.e. the error rate) for classification depends on not only the margin of positive/negative samples but also the radius of the enclosing ball of all samples, and this is more critical for joint learning of feature representation and classifier. Inspired by these works, we incorporate a radius-margin bound as a regularizer into our deep model, and demonstrate better generalization performance compared to the softmax or SVM classifier. More detailed discussion will be presented Section 3.3.}
%
%capability on learning with small scale data. Unfortunately, these approaches are discussed mainly using the fixed features, and they are infeasible to directly work with the feature learning for the reason that the shifty features can affect the classification margin. For example, simply multiplying a constant to the feature vector can enlarge the margin between the positive and negative samples, but obviously it will not lead to a better result. In this work, we solve this problem by introducing a radius bound into the max-margin method, inspired by the minimum enclosing ball (MEB) learning \cite{R98}. Specifically, we calculate the radius of the enclosing ball of all samples in the feature space, and fix the radius while maximizing the classification margin. Intuitively, this radius bound constrain the training samples not to be diverged, and thus ensures the availability of maximizing classification margin.

Training our deep structured model is nontrivial, as it needs to jointly optimize three components: (i) the activity decomposition, (ii) the classifier upon the generated features, and (iii) the neural networks. Seeking the global optimum for such a model is extremely intractable due to the non-convexity, and we consider an approximate solution by iteratively optimizing these components for a local convergence. In each iteration, the learning algorithm performs the following three steps.
\begin{enumerate}
    \item We compute the optimal latent variables (i.e. sub-activity decompositions) for all training activities, and their feature vectors are then specified.
    \item Based on the generated features, we optimize the classification margin of all training examples under the fixed radius bound.
    \item We learn the parameters of the CNNs using the traditional backward propagation, which will lead to the decrease of the radius.
\end{enumerate}

The main contributions of this work are several folds. First, we present a novel deep neural network model to handle various challenges in 3D human activity recognition, and demonstrate superior performance over state-of-the-art approaches under several challenging scenarios. { Second, our deep model incorporates latent temporal structure to account for large temporal variations of diverse human activities.} To the best of our knowledge, this is a novel contribution to the literature of deep learning. Third, we unify the radius-margin method with the feature learning in a principled way, providing a very general framework for many classification tasks. In addition, we construct a new database of RGB-D data, which includes 1180 instances of human activities in 20 categories.

The remainder of the paper is organized as follows. Section 2 presents a review of related work. Then we present our deep model in Section 3 and 4, followed by a description of model learning algorithm in Section 5. Section 6 discusses the procedure of activity recognition using our model. The experimental results, comparisons and component analysis are exhibited in Section 6. Section 7 concludes this paper.

%To summarize, the novelties of this paper include:(1) A deep model with structural alternatives for activity recognition; (2)An  EM-type optimization algorithm for deep learning; (3) Transfer learning to borrow strength from 2D data to 3D data.

\section{Related Work}

Many works on human action/activity recognition mainly focus on designing robust and descriptive features~\citep{DSTIP,diMM13,HON4D,FG2013,BagSIFT,DMMM12, 3dsiftMM07}. For example, Xia and Aggarwal~\citep{DSTIP} extracted spatio-temporal interest points from depth videos (DSTIP) and developed a depth cuboid similarity feature (DCSF) to model human activities. Oreifej and Liu~\citep{HON4D} proposed to capture spatio-temporal changes of activities by using a histogram of oriented 4D surface normals (HON4D). Most of these methods, however, overlooked detailed spatio-temporal structure information, and limited in periodic activities.

Several compositional part-based approaches have been studied for complex scenarios and achieved substantial progresses \citep{WangPAMI2011,AlanCVPR2013,MMPose2013,CVPR12PoseObject,JCorsoCVPR2012,WuYingCVPR2012,shuichengMM2011}, and they represent an activity with the deformable parts and contextual relations. For instance, Wang et al.~\citep{WangPAMI2011} recognized human activities in common videos by training the hidden conditional random fields in a max-margin framework. For activity recognition in RGB-D data, Packer et al.~\citep{CVPR12PoseObject} employed the latent structural SVM to train the model with part-based pose trajectories and object manipulations. An ensemble model of actionlets were studied in \citep{WuYingCVPR2012} to represent 3D human activities with a new feature called local occupancy pattern (LOP).  To handle more complicated activities with large temporal variations, some improved models ~\citep{FeifeiCVPR2012,WuYingICCV2013,SinisaICCV2011} discovered temporal structures of activities by localizing sequential actions. For example, Wang and Wu~\citep{WuYingICCV2013} proposed to solve the temporal alignment of actions by maximum margin temporal warping. Tang et al.~\citep{FeifeiCVPR2012} captured the latent temporal structures of 2D activities based on the variable-duration hidden Markov model. Koppula and Saxena~\citep{SaxenaICML2013} applied the Conditional Random Fields to model the sub-activities and affordances of the objects for 3D activity recognition.

%handled the action recognition by modeling both pose trajectories and object manipulations with a latent structural SVM
%
%In the scenario of depth video, Packer et al.~\citep{CVPR12PoseObject} handled the action recognition by modeling both pose trajectories and object manipulations with a latent structural SVM. Wang et al.~\citep{WuYingCVPR2012} developed an actionlet ensemble model and a novel feature called local occupancy pattern(LOP) to capture the intra-class variance in 3D action. However, these methods only address small time period action recognition, where temporal segmentation matters a little.

Recently, the And-Or graph representations are introduced as extensions of the part-based models~\citep{AOGZhu2006,AOGICCV2011,LiangMM2013,JVTPMM14,LinGrammar}, and produce very competitive performance to deal with large data variations. These models incorporate not only the hierarchical decompositions, but also the explicit structural alternatives (e.g. the different ways of compositions). Zhu and Mumford~\citep{AOGZhu2006} first explored the And-Or graph models for image parsing. Pei et al.~\citep{AOGICCV2011} then introduced the models for video event understanding, but their approach required elaborate annotations. Liang et al.~\citep{LiangMM2013} proposed to train the spatio-temporal And-Or graph model using a non-convex formulation, which is discriminatively trained from weakly
annotated training data. However, the above mentioned models rely on the hand-crafted features, and their discriminative capacities are not optimized for 3D human activity recognition.

In the mean time, the past few years have seen a resurgence of research in the design of deep neutral networks, and impressive progresses have been made on learning image features from raw data~\citep{Hinton06,ImagenetNIPS2012, DSPACV13}. To address human action recognition from videos, Ji et al.~\citep{3DCNNPAMI} developed a novel deep architecture of convolutional networks, where they extracted features from both spatial and temporal dimensions. Luo et al. \cite{SwitchDNNs} proposed to incorporate a new Switchable Restricted Boltzmann Machine (SRBM) to explicitly model the complex mixture of visual appearance for pedestrian detection, and train their model using an EM-type interative algorithm. Amer and Todorovic~\citep{SinisaSPNCVPR2012} applied Sum Product Networks (SPNs) to model human activities based on variable primitive actions. Our deep model is partially motivated by these works, and we target on an more flexible and powerful solution by jointly considering the latent structure embedding, feature learning, and radius-margin classification.

{ Recently, recurrent neural networks (RNN) has been used for activity recognition due to its capability in modeling complex temporal dynamics. Donahue et al.~\cite{LRNN} presented a long-term recurrent convolutional network (LRCN) architecture to integrates CNN and RNN into an unified model, and achieved promising results in a number of vision tasks. Rohrbach et al.~\cite{RNN} further improved LRCN by adding a pooling layer and had shown its potentials in video description. The main difference between RNN models and ours is that their models exploit several types of neural gates and memory cells to learn temporal dynamics implicitly, while our deep structured model explicitly accounts for temporal variations of human activities by inferring latent variables. Speficially, compared with these RNN models, our model has the following advantages. First, the temporal composition is explicitly captured by our model, giving rise to a better interpretability, i.e. the semantic correspondence of video segments and sub-activities. Second, as some recent works report~\cite{bayer2014a}, the RNN models may have problems on using common dropout tricks and this limitation would influence the performances.  Moreover, the integration with explicit regularization approaches (e.g. the radius-margin bound) is also an important superiority of our model.}

%Although deep models have impressive results in 2D action and activity recognition, there are few studies on 3D action.

%The main reason is that the lack of data leads to the difficulties in learning complex deep models. The very recent study by Girshick et al.~\citep{rbgCVPR2014} introduced an object recognition method, which pre-trains Convolutional Neural Network on a large dataset and transfers its learned parameters to solve the problem in a smaller one. It inspires us to do transfer learning by borrowing the data strength from 2D video data to train our 3D deep model for activity recognition in RGB-D data.

\section{Deep Structured Model}

In this section, we introduce the main components of our deep structured model, including the { spatio-temporal} CNNs, the latent structure of activity decomposition, and the radius-margin bound for classification.

\subsection{Spatio-temporal CNNs}

%Our deep model is presented as a spatio-temporal convolutional neural network, as shown in Figure~\ref{fig:Architecture}. To model the complex human activities, it comprises of $M$ network cliques, which jointly conduct the final output. We define a clique as a subpart of the network stacked up for several layers. In particular, each clique extracts features from one decomposed video segment associated to one separated action from the complete activity, and an illustration is highlighted in Figure~\ref{fig:Architecture}. Specifically, for each clique, two 3D convolutional layers are first built upon the raw input (i.e., grayscale and depth data), which consists with at most $m$ video frames, and then followed by one 2D convolutional layer. Note that a max-pooling operator is applied on each 3D convolutional layer making our model robust to local body deformations and surrounding noises. Afterwards, the convolution results generated by different cliques are merged and concatenated into a long feature vector, upon which we build two full connection layers to associate with the activity labels.  In the following, we introduce the detailed definitions for these components of our model.

We propose an architecture of spatio-temporal convolutional neural networks (CNNs), as Figure~\ref{fig:Architecture} illustrates. In the input layer, the activity video is decomposed into $M$ video segments, where each segment associates to one separated sub-activity. Accordingly, { the proposed architecture consists of $M$ sub-networks to extract features from the corresponding decomposed video segments}, respectively. Our spatio-temporal CNNs involve both 3D and 2D convolutional layers. The 3D convolutional layer extracts spatio-temporal features for jointly capturing appearance and motion information, and is followed by a max-pooling operator to improve the robustness against local deformations and noise.
As shown in Figure~\ref{fig:Architecture}, { each sub-network (highlighted by the dashed box) is stacked up by two 3D convolutional layers and one 2D convolutional layer}. { For the input to each sub-network, the number of frames is very small (e.g. 9). After two layers of 3D convolution followed with max-pooling, the temporal dimension for each set of feature maps is too small to perform 3D convolution. Thus, we stack a 2D convolutional layer upon the two 3D convolutional layers.}{ The outputs from different sub-networks are merged to be fed to one fully connected layer that generates the final feature vector of the input video.}

\begin{figure}[!htb]
\centering
\includegraphics[width=\columnwidth]{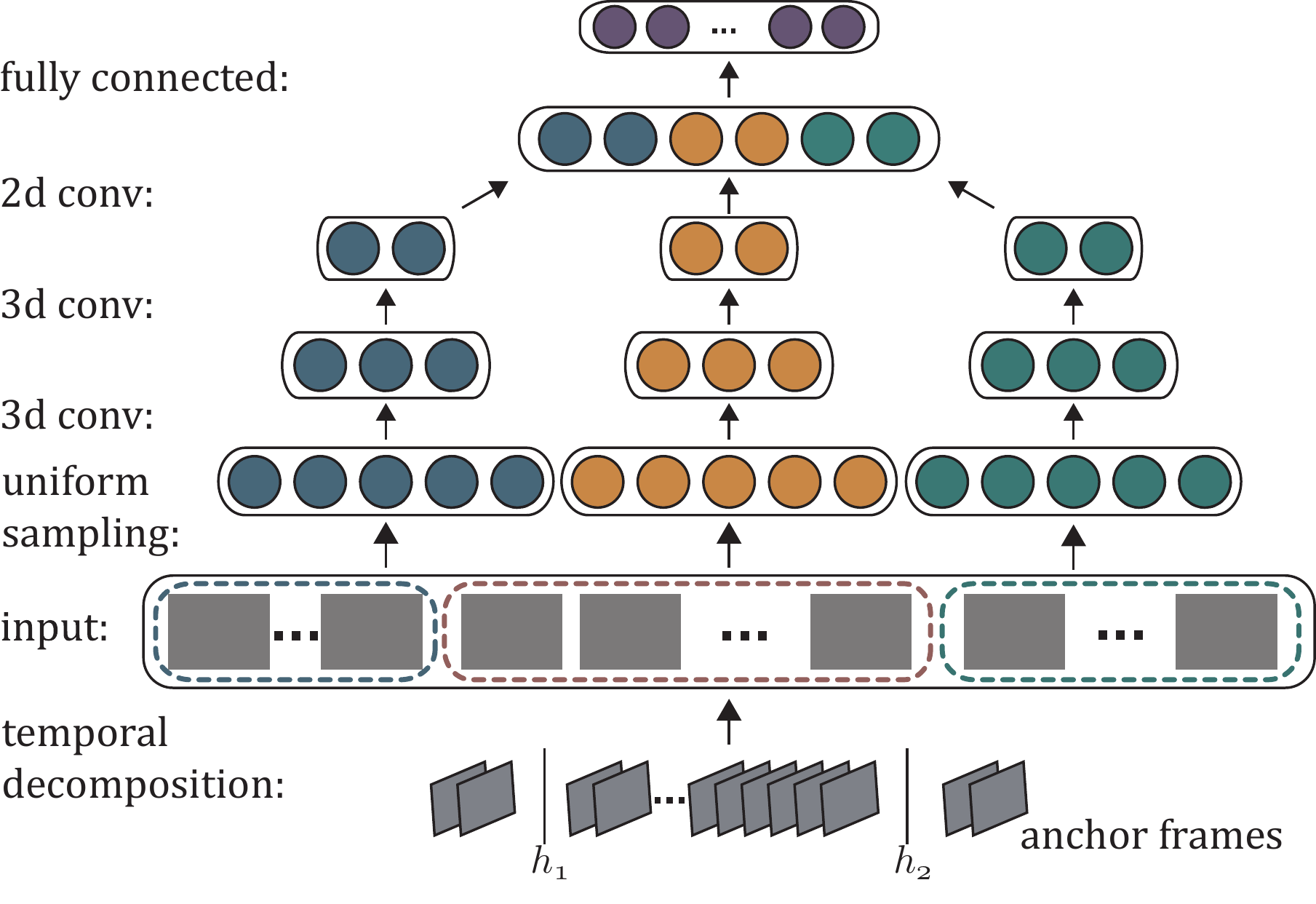}
\caption{ { Illustration of incorporating latent structure into the deep model. Different sub-networks are denoted by different colors. }}\label{fig:latent_structure}
\end{figure}

\subsection{Latent Temporal Structure}

Unlike the traditional deep learning methods with the fixed architectures, we incorporate latent structure into the deep model to flexibly adapt to the input video during inference and learning. { To address the large temporal variation of human activities, we assume the input video is temporally divided into a number $M$ of segments, corresponding to the sub-activities. We associate the CNNs with the video segmentation by feeding each segmented part into a sub-network as Figure~\ref{fig:Architecture} illustrates. Next, according to the way of video segmentation (i.e. decomposition of sub-activities), we manipulate the CNNs by inputting sampled video frames. 

Specifically, we index each video segment by its starting anchor frame $s_j$ and its temporal length (i.e. the number of frames) $t_j$ for each sub-network, which must take $m$ video frames as the input. Note that when $t_j \neq m$, a uniform sampling is performed to extract $m$ key frames. Thus, for all video segments, we denote the indexes of starting anchor frames as $(s_1,...,s_M)$ and their temporal lengths as $(t_1,...,t_M)$, which are regarded as the latent variables in our model,  $h=(s_1,...,s_M, t_1,...,t_M)$. These latent variables specifying the segmentation will be adaptively estimated for different input videos. Figure~\ref{fig:latent_structure} shows an intuitive example of our structured deep model, where the input video are segmented into three sections corresponding to the three sub-networks in our deep architecture. In this way, the configuration of the CNNs are dynamically adjusted together with searching for the appropriate latent variables of input videos.}

Given the parameters of CNNs $\omega$ and the input video $x_i$ with its latent variables $h_i$, the generated feature of $x_i$ can be represented as $\phi(x_i; \omega, h_i)$.

\begin{figure*}[!htb]
\centering
\includegraphics[width=\textwidth]{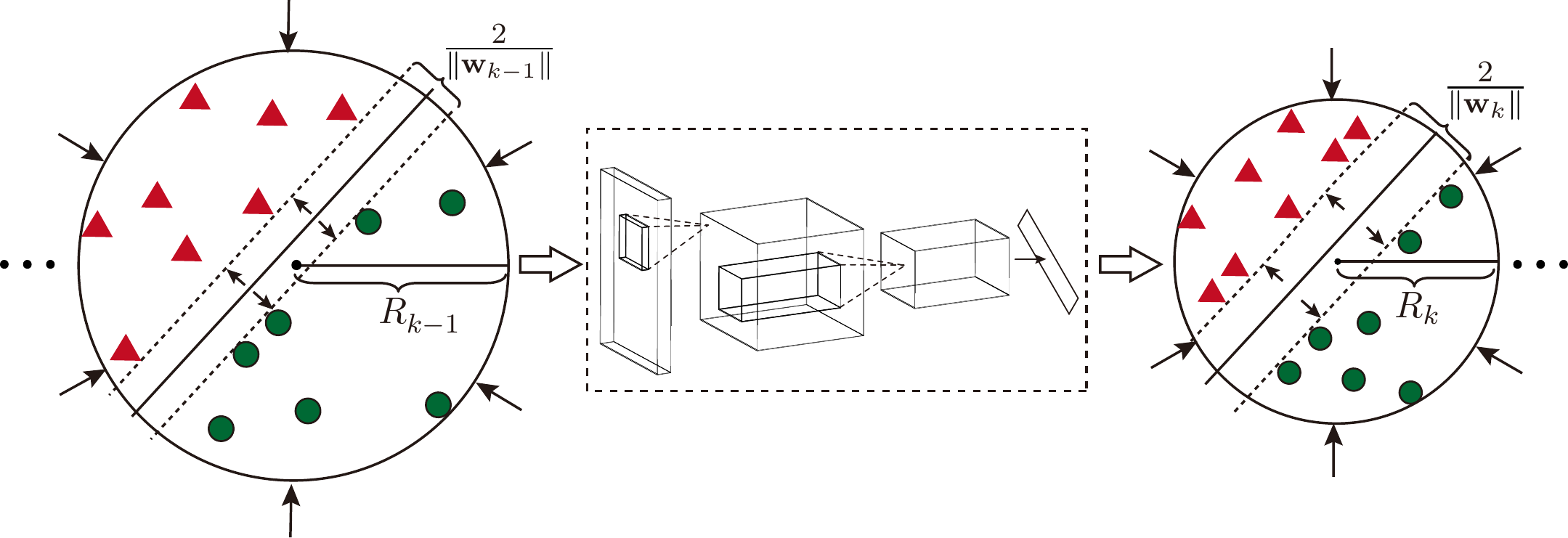}
\caption{ Illustration of our deep model with the radius-margin bound. To improve the generalization performance for classification, we propose to integrate the radius-margin bound as a regularizer with the feature learning. Intuitively, together with optimizing the max-margin parameters $({\bf w}, b)$, we shrink the radius $R$ of the minimum enclosing ball (MEB) of the training data that distribute in the feature space generated by the CNNs. The resulting classifier with the regularizer shows better generalization performance compared to the traditional softmax output. }\label{fig:classifier}
\end{figure*}

%{
%Recently, combining Long and Short Term Memory network (LSTM) with a deep hierarchical visual feature extractor (e.g, CNN) for activity recognition has achieved promising results~\cite{LRNN, RNN}. In addition to the latent variable, the LSTM includes an input gate, forget gate, output gate, input modulation gate, and memory cell. These additional components enable the LSTM to learn extremely complex and long-term temporal dynamics implicitly.
%
%Though both these methods and our deep model contain latent variables to model temporal information implicitly and allow end-to-end fine-tuning, there exists a major difference: Our deep model utilizes latent variables to model the temporal composition of human activities (global temporal information), and applies 3D convolutional layers to extract spatio-temporal features (local temporal information). In such a global and local consideration, our deep model can be partially clarified and carefully analyzed for activity recognition. But for LSTM, temporal representation is totally invisible and difficult to understand due to its latent variables with varying connections within the memory cells. Thus, our proposed model is less flexible, but has a more explanatory power than LSTM used in \cite{LRNN, RNN}.
%}

\subsection{Deep Model with Relaxed Radius-Margin Bound}

The large amount of training data is crucial for the success of many deep learning models. Given sufficient training data, the effectiveness of applying the softmax classifier with CNNs has been validated for image classification~\citep{ImagenetNIPS2012}. However, for 3D human activity recognition, the available training data are usually less what we expected. For example, the CAD-120 dataset~\cite{CADIJRR2013} consists of only 120 RGB-D sequences of 10 categories. { Under this scenario, though parameter pre-training and dropout are available, the model training often suffers from the over-fitting issue. Hence, we consider introducing a more effective classifier together with regularizer to improve the generalization performance of the deep model.}

In supervised learning, Support Vector Machine (SVM), also known as the max-margin classifier, is theoretically sound and generally can achieve promising performance compared with the alternative linear classifiers. In the deep learning research, the combination of SVM and CNNs has been exploited~\citep{Huang06} and obtained excellent results in object detection~\citep{rcnn2014}. Motivated by these approaches, we impose a max-margin classifier $({\bf w}, b)$ upon the feature generated by the spatio-temporal CNNs for human activity recognition.

{ As a max-margin classifier, standard SVM adopts $\| {\bf w} \|^2$, the reciprocal of the squared margin $\gamma^2$, as the regularizer. However, the generalization error bound of SVM depends on the radius-margin ratio $R^2/\gamma^2$, where $R$ is the radius of the minimum enclosing ball (MEB) of the training data~\citep{R98}.  When the feature space is fixed, the radius $R$ is constant and can thus be ignored. However, in our approach, the radius $R$ is determined by the MEB of the training data in the feature space generate by the CNNs. Under this scenario, the model has the risk that the margin can be increased by simply expanding the MEB of the training data in the feature space. For example, simply multiplying a constant to the feature vector can enlarge the margin between the positive and negative samples, but obviously it will not really work for better classification. To overcome this problem, we incorporate the radius-margin bound together with the feature learning, as Figure~\ref{fig:classifier} illustrates}. In particular, we impose a max-margin classifier with radius information upon the feature generated by the fully connected layer of the spatio-temporal CNNs. The optimization tends to maximize the margin while shrinking the MEB of the training data in the feature space, and we thus obtain a tighter error bound.

Suppose there are a set of $N$ training samples $(X, Y)$ = \{$(x_1,y_1)$, ... , $(x_N,y_N)$\}, where $x_i$ is the video, $y \in \{1,...,C\}$ represents the category labels and $C$ is the number of activity categories. We extract the feature for each $x_i$ by the spatio-temporal CNNs, $\phi(x_i; \omega, h_i)$, where $h_i$ refers to the latent variables. By adopting the squared hinge loss and the radius-margin bound, we define the following loss function $L_0$ of our model:

\begin{equation}
\label{equ:svm}
\begin{split}
L_0 = &\overbrace{\frac{1}{2} \Arrowvert {\bf w} \Arrowvert^2 R_{\phi}^2}^{Radius-margin~Ratio} \\
&+ \lambda \sum_{i=1}^N \max \bigg( 0, 1 - \big({\bf w}^T\phi(x_i; \omega, h_i) + b \big)y_i \bigg)^2,
\end{split}
\end{equation}
where $\lambda$ is the trade-off parameter, $1/\Arrowvert {\bf w} \Arrowvert$ denotes the margin of the separating hyperplane, $b$ denotes the bias, and $R_{\phi}$ denotes the radius of the MEB of the training data $\phi(X, \omega,H)$ = $\{\phi(x_1;\omega,h_1),...,\phi(x_N;\omega,h_N)\}$ in the CNNs' feature space. Formally, the radius $R_{\phi}$ is defined as~\citep{R02, R98},

\begin{equation}
\label{equ:defR}
\begin{gathered}
 R_{\phi}^2 = \min_{R, \phi_0} R^2, s.t. \| \phi(x_i;\omega,h_i) - \phi_0 \|^2 \leq R^2, \forall i.
\end{gathered}
\end{equation}

%When combining CNNs with SVM, the standard max-margin classifier ignores the radius of the MEB of $\phi(X, \omega,H)$, may leads to higher error bound and results in poor classification performance. As illustrated in Figure~\ref{fig:classifier}, the introduction of radius-margin bound can guide the learning of CNN model parameters to make $\phi(X, \omega, H)$ to have a small radius while being classified with max-margin, and thus can improve the classification and generalization performance.

The radius $R_{\phi}$ is implicitly defined by both the training data and the model parameters, making that: (i) the model in Eq. (\ref{equ:svm}) is highly nonconvex, (ii) the derivative of $R_{\phi}$ with respect to $\omega$ is hard to compute, and (iii) the problem is difficult to solve using the stochastic gradient descent (SGD) method. Motivated by the radius-margin based SVM \citep{R-SVM,RSVMICML13}, we investigate the relaxed form to replace the original definition of $R_{\phi}$ in Eq. (\ref{equ:defR}). In particular, we introduce the maximum pairwise distance $\tilde{R}_{\phi}$ over all the training samples in the feature space, as

\begin{equation}
\label{equ:def_tildeR}
\begin{gathered}
 \tilde{R}_{\phi}^2 = \max_{i,j} \| \phi(x_i;\omega,h_i) - \phi(x_j;\omega,h_j) \|^2.
\end{gathered}
\end{equation}
Do and Kalousis \cite{RSVMICML13} proved that $R_{\phi}$ could be well bounded by $\tilde{R}_{\phi}$ with the following lemma,
\begin{lemma}\label{lemma:radius_bound}
$$ \tilde{R}_{\phi} \leq R_{\phi} \leq \frac{1+\sqrt{3}}{2} \tilde{R}_{\phi}.$$
\end{lemma}

This above Lemma guarantees that the true radius $R_{\phi}$ can be well approximated by $\tilde{R}_{\phi}$. With the proper parameter $\eta$, the optimal solution for minimizing the radius-margin ratio $\| {\bf w} \|^2 {R}_{\phi}^2$ is the same with that for minimizing the radius-margin sum $\| {\bf w} \|^2 + \eta {R}_{\phi}^2$~\citep{RSVMICML13}. Thus, by approximating ${R}_{\phi}^2$ with ${\tilde{R}}_{\phi}^2$ and replacing the radius-margin ratio with the radius-margin sum, we suggest the following deep model with the relaxed radius-margin bound,

\begin{equation}
\label{equ:svm_L1}
\begin{split}
L_1 = \frac{1}{2} \Arrowvert {\bf w} \Arrowvert^2 + \max_{i,j} \| \phi(x_i;\omega,h_i) - \phi(x_j;\omega,h_j) \|^2 \\
+ \lambda \sum_{i=1}^N \max \bigg( 0, 1 - \big({\bf w}^T\phi(x_i; \omega, h_i) + b \big)y_i \bigg)^2.
\end{split}
\end{equation}

However, the { first} max operator in Eq. (\ref{equ:svm_L1}) is non-smooth and defined over all pairs of training samples, and it is thus unsuitable for using the mini-batch-based SGD optimization method. { In the following, we first use the softmax function  to avoid the non-smoothness of the max operator, and then further relax the radius to avoid the definition over all pairs of training samples. More specifically, we first transfer the max operator into a softmax form, resulting the following model,}

\begin{equation}
\label{equ:svm_L2}
\begin{split}
L_2 = \frac{1}{2} \Arrowvert {\bf w} \Arrowvert^2 + \eta \sum_{i,j} \kappa_{ij} \| \phi(x_i;\omega,h_i) - \phi(x_j;\omega,h_j) \|^2 \\
+ \lambda \sum_{i=1}^N \max \bigg( 0, 1 - \big({\bf w}^T\phi(x_i; \omega, h_i) + b \big)y_i \bigg)^2.
\end{split}
\end{equation}
with
\begin{equation}
\label{equ:def_weight}
\begin{gathered}
 \kappa_{ij} = \frac{\exp (\alpha \| \phi(x_i;\omega,h_i) - \phi(x_j;\omega,h_j) \|^2)}{\sum_{ij} \exp (\alpha \| \phi(x_i;\omega,h_i) - \phi(x_j;\omega,h_j) \|^2)},
\end{gathered}
\end{equation}
{ where $\kappa_{ij}$ is a coefficient measuring the correlation of the two samples and $\alpha \geq 0$ is the parameter to control the approximation degree to the hard max operator.} { When $\alpha$ is infinite, the approximation in Eq. (\ref{equ:svm_L2}) becomes the model in Eq. (\ref{equ:svm_L1}).} Specifically, when $\alpha = 0$, there is $\kappa_{ij} = 1/{N^2}$, and the relaxed loss function can be reformulated as:

\begin{equation}
\label{equ:svm_L3}
\begin{split}
L_3 = & \frac{1}{2} \Arrowvert {\bf w} \Arrowvert^2 + {2\eta} \sum_{i} \| \phi(x_i;\omega,h_i) - \bar{\phi}_{\omega} \|^2 \\
&+ \lambda \sum_{i=1}^N \max \bigg( 0, 1 - \big({\bf w}^T\phi(x_i; \omega, h_i) + b \big)y_i \bigg)^2.
\end{split}
\end{equation}
with
\begin{equation}
\label{equ:def_mean}
\begin{gathered}
 \bar{\phi}_{\omega} = \frac{1}{N}{\sum_{i} \phi(x_i;\omega,h_i)}.
\end{gathered}
\end{equation}

The optimization objectives in Eq. (\ref{equ:svm_L2}) and (\ref{equ:svm_L3}) are two relaxed losses of our deep model with the strict radius-margin bound in Eq. (\ref{equ:svm}). { In this work, we focus on the objective in Eq. (\ref{equ:svm_L3}) for the model training.} The learning algorithm will be discussed in Section \ref{sec:learning}.

\section{Implementation}

In this section, we first explain the implementation { that makes our model adaptive to alterable temporal structure, and then describe the detailed setting of our deep architecture.}

\subsection{Latent Temporal Structure}

{ During our learning and inference procedures, we search for the appropriate latent variables that determine the temporal decomposition of the input video (i.e. the decomposition of activities).}  There are two parameters related to the latent variables in our model: the number $M$ of video segments and the temporal length $m$ of each segment. Note that the sub-activities decomposed by our model have no precise definition given a complex activity, i.e. actions can be ambiguous depending on the considering temporal scale.

To incorporate the latent temporal structure, we associate the latent variables with the neurons (i.e. convolutional responses) in the bottom layer in the spatio-temporal CNNs.

The choice of the number of segments $M$ is important to the performance of 3D human activity recognition. The model with a small $M$ could be less expressive to handle temporal variations, while a large $M$ could lead to over-fitting due to high complexity. Furthermore, { when $M = 1$, the model latent structure would be disabled, and our architecture degenerates to the conventional 3D-CNNs~\cite{3DCNNPAMI}}. By referring to the setting of the number of parts for the deformable part-based model~\cite{lsvm-pami} in object detection, the value $M$ can be set by the cross validation on a small set. In all our experiments, we set $M$ = 4.

{
Considering that the number of frames of the input videos are diverse, we develop a process to normalize the inputs by two-step sampling in the learning and inference procedure. First, we sample $30$ anchor frames uniformly from the input video. Based on these anchor frames, we search for all of the possible non-overlapped temporal segmentations, and the anchor frame segmentation corresponds to the segmentation of the input video. Then, from each video segment (indicating a sub-activity) we uniformly sample $m$ frames to feed the neural networks, and in our experiments we set $m = 9$. In addition, we reject the possible segmentations that cannot offer $m$ frames for any video segment.

For an input video, the possibility of temporal structure variations is $115$ in our experiments (i.e. the possible enumeration numbers of anchor frame segmentation).
}

\begin{figure}[!ht]
\centering
\includegraphics[width=3.2in]{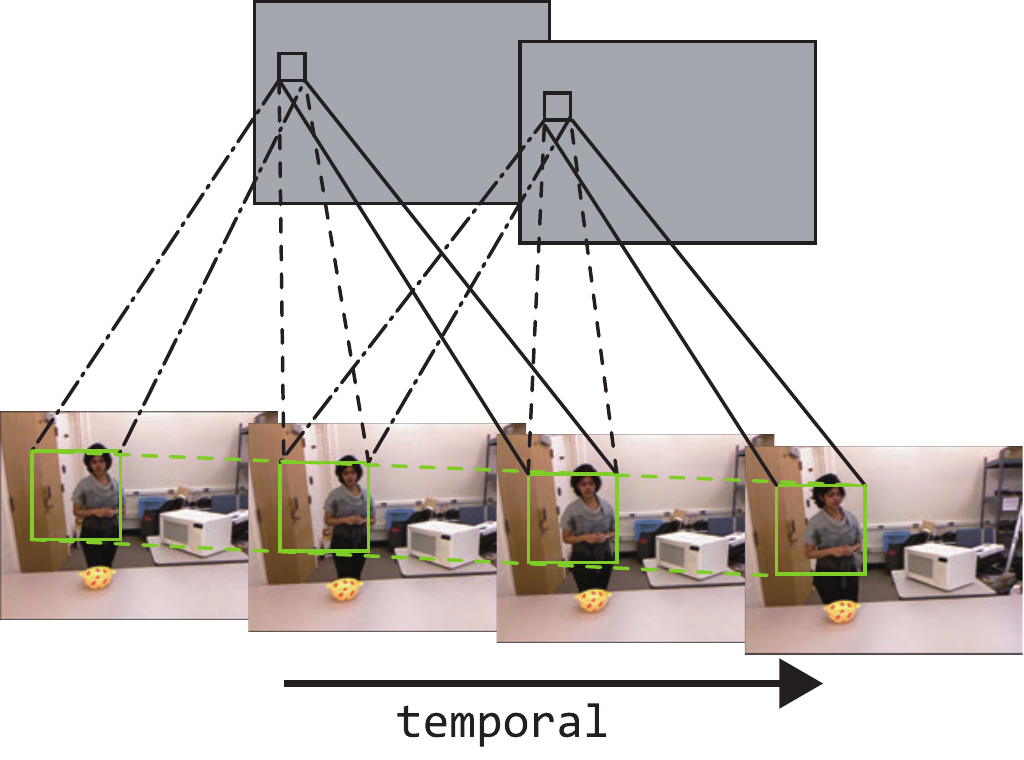}
\caption{Illustration of the 3D convolution across both spatial and temporal domains. In this example, the temporal dimension of the 3D kernel is 3, and each feature map is thus obtained by performing 3D convolutions across 3 adjacent frames.}\label{fig:conv3d}
\end{figure}

\subsection{Architecture of Deep Neural Networks}

The proposed spatio-temporal CNN architecture is constructed by stacking up two 3D convolution layers, one 2D convolution layer and one fully connected layer, and the max-pooling operator is deployed after each 3D convolutional layer. In the following, we introduce the definitions and implementations of these components in our model.

\textbf{3D Convolutional Layer.} The 3D convolution operation is adopted to perform convolutions spanning over both spatial and temporal dimensions for the characterization of both appearance and motion features~\citep{3DCNNPAMI}. Suppose ${\bf p}$ is the input video segment with the width $w$, the height $h$, and the number of frames $m$, $\omega$ is the 3D convolutional kernel with the the width $w^{\prime}$, height $h^{\prime}$, and temporal length $m^{\prime}$. As shown in Figure~\ref{fig:conv3d}, a feature map ${\bf v}$ can be obtained by performing 3D convolutions from the $s$th to the $(s+m^{\prime}-1)$th frames, where the response for the position $(x,y,s)$ in the feature map is defined as,

\begin{eqnarray} \label{eq:3Dconv}
v_{xys} = \tanh(b + \sum_{i=0}^{k^{\prime}-1} \sum_{j=0}^{h^{\prime}-1} \sum_{k=0}^{m^{\prime}-1} \omega_{ijk} \cdot p_{(x+i)(y+j)(s+k)}),
\end{eqnarray}
where $p_{(x+i)(y+j)(s+k)}$ denotes the pixel value of the input video ${\bf p}$ at position $(x+i,y+j)$ in the $(s+k)$th frame, $\omega_{ijk}$ denotes the value of the convolutional kernel $\omega$ at the position $(i,j,k)$, $b$ stands for the bias, and $\tanh$ denotes the hyperbolic tangent function. Thus, given ${\bf p}$ and $\omega$, $m - m^{\prime} + 1$ feature maps can be obtained, each with size of $(w - w^{\prime}+ 1, h - h^{\prime}+ 1)$.

Based on the 3D convolution operation, 3D convolution layer is designed for spatio-temporal feature extraction by considering three issues:

\begin{itemize}
    \item {\emph{Number of convolutional kernels.}} The feature maps generated by one convolutional kernel are limited in capturing appearance and motion information. To generate more types of features, several kernels are employed in each convolutional layer. We define the number of 3D convolutional kernels in the first layer as $c_1$. After the first 3D convolutions, we obtain $c_1$ sets of $m - m^{\prime} + 1$ feature maps. Then we use 3D convolutional kernels on the $c_1$ sets of feature maps, and obtain $c_1 \times c_2$ sets of feature maps after the second 3D convolution layer.

    \item {\emph{Decompositional convolutional networks.}} { Our deep model consists of $M$ sub-networks, and the input video segment to each sub-network involves $m$ frames (the later frames might be unavailable). { In the proposed architecture, all of the sub-networks use the same structure but each one has its own convolutional kernels, as we assume that each temporally decomposed sub-activity has its distinct features in terms of discriminative classification}. For example, the kernels belonging to the first sub-network are only deployed to perform convolutions on the first temporal video segment. }.

    \item {\emph{Application to gray-depth video.}} The RGB images are first converted to the gray-level images, and the gray-depth video is then adopted as the input to the neural networks. The 3D convolutional kernels in the first layer are respectively applied for both the gray channel and the depth channel in the video, and the convolution results from these two channels are further aggregated to produce the feature maps. Note that the dimensions of the features remain the same as from only one channel.

\end{itemize}

In our implementation, the input frame is scaled with the height $h = 80$ and width $w = 60$. In the first 3D convolution layer, the number of 3D convolutional kernels is $c_1=7$, and the size of the kernel is $w^{\prime} \times  h^{\prime} \times m^{\prime}$ = $9 \times 7 \times 3$. In the second layer, the number of 3D convolutional kernels is $c_2=5$, and the size of the kernel is $w^{\prime} \times  h^{\prime} \times m^{\prime}$ = $7 \times 7 \times 3$. Thus, we have $7$ sets of feature maps after the first 3D convolution layer, and obtain $7 \times 5$ sets of feature maps after the second 3D convolution layer.

\textbf{Max-pooling Operator.} After each 3D convolution, the max-pooling operation is introduced to enhance the deformation and shift invariance~\citep{ImagenetNIPS2012,KaiYuSC}. Given a feature map with the size of $a_1 \times a_2$, a $d_1 \times d_2$ max-pooling operator is performed by taking the maximum of every non-overlapping $d_1 \times d_2$ sub-regions of the feature map, resulting in an  $a_1/d_1 \times a_2/d_2$ pooled feature map. In our implementation, $3 \times 3$ max-pooling operator was applied after every 3D convolution layers. After two layers of 3D convolution and max-pooling, for each sub-network, we have $7 \times 5$ sets of $6 \times 4 \times 5$ feature maps.

\textbf{2D Convolutional Layer.} After two layers of 3D convolution followed with max-pooling, 2D convolution is employed to further extract higher-level complex features. The 2D convolution can be viewed as a special case of 3D convolution with $m^{\prime}  = 1$, which is defined as

\begin{eqnarray} \label{eq:2Dconv}
v_{xy} = \tanh(b + \sum_{i=0}^{k^{\prime}-1} \sum_{j=0}^{h^{\prime}-1} \omega_{ij} \cdot p_{(x+i)(y+j)}),
\end{eqnarray}
where $p_{(x+i)(y+j)}$ denotes the pixel value of the feature map ${\bf p}$ at position $(x+i,y+j)$, $\omega_{ij}$ denotes the value of the convolutional kernel $\omega$ at the position $(i,j)$, and $b$ denotes the bias. In the 2D convolution layer, suppose the number of 2D convolutional kernels is $c_3$, $c_1 \times c_2 \times c_3$ sets of new feature maps are obtained by performing 2D convolutions on $c_1 \times c_2 $ sets of feature maps generated by the the second 3D convolution layer.

In our implementation, the number of 2D convolutional kernels is set as $c_3=4$ with the kernel size $6 \times 4$. Hence for each sub-network we can obtain $700$ feature maps with size $1 \times 1$.

\textbf{Fully Connected Layer.} There is only one fully connected layer with 64 neurons in our architecture. All these neurons connect to a vector of $ 700 \times 4 = 2800 $ dimensions, which is generated by concatenating the feature maps from all of the sub-networks. The margin-based classifier is defined based on the output of the fully connected layer, where we adopt the squared hinge loss to predict the activity categories as
\begin{equation}
\theta(z) = \arg \max_i ( w_{i}^{T} z + b_i )
\end{equation}
where $z$ is the 64-dimensional vector from the fully connected layer, and $\{ w_i, b_i \}$ denotes the weight and bias connected to the $i$-th activity category.

{ {\bf Dropout trick.} Since our deep architecture contains a large number of parameters (i.e. 179200 weighs at the fully-connected layer), we apply the standard dropout approach during the model training to alleviating the over-fitting problem. According to the recent reports~\cite{dropout}, the dropout method is capable of effectively improving the generalization power of neural network models by randomly turning off the neurons in the learning. Specifically, we set turning-off probability rate is $0.6$ for each neuron at the fully-connected layer in each learning iteration, and this dropout approach is applied by default in every experiment.
}

%----------------------------------------------------------------------------------------
\section{Learning Algorithm} \label{sec:learning}

{ The proposed deep structured model involves three components to be optimized: (i) the latent variables $H$ that manipulate the activity decomposition, (ii) the margin-based classifier \{${\bf w}, b$\}, and (iii) the CNNs' parameters $\omega$.} The latent variables are not continuous and need to be estimated adaptively for different input videos, making the standard back propagation algorithm~\citep{CNN1990} unsuitable for our deep model. In this section, we present a joint component learning algorithm that iteratively optimizes the three components. Moreover, to overcome the problem of insufficient 3D data, we propose to borrow the large amount of 2D videos to pre-train the CNNs' parameters in advance.

\begin{figure*}[!htb]
\centering
\includegraphics[width=1.6 \columnwidth]{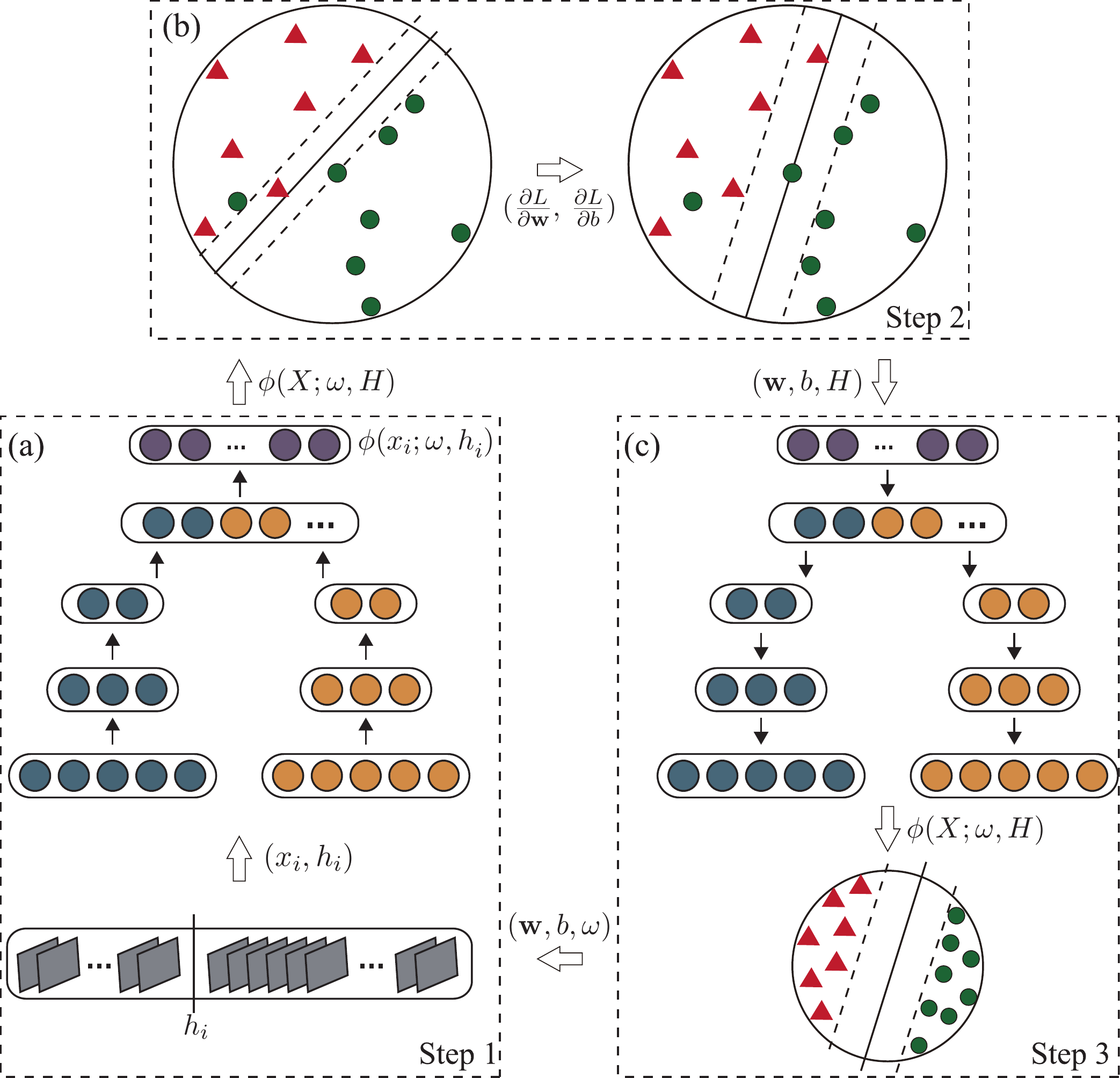}
\caption{ Illustration for our joint component learning algorithm. It iteratively performs with the three steps: (a) Given the classification parameters \{${\bf w}, b$\} and the CNNs' parameters $\omega$, we estimate the latent variables $h_i$ for each video and generate the corresponding feature $\phi(x_i; \omega, h_i)$; (b) Given the updated features $\phi(X; \omega, H)$ for all training examples, the classifier \{${\bf w}, b$\} is updated via SGD ; (c) Given \{${\bf w}, b$\} and $H$, back propagation updates the CNNs' parameters $\omega$. }\label{fig:learning}
\end{figure*}

\subsection{Joint Component Learning}

Denote $(X, Y)$ = \{$(x_1,y_1)$, ... , $(x_N,y_N)$\} as the training set with $N$ examples, where $x_i$ is the video, $y_i \in \{1,...,C\}$ denotes the activity category. Denote ${H} = \{h_1,...,h_N\}$ as the set of latent variables for all training examples. The model parameters to be optimized can be divided into three groups, i.e. ${H}$, \{${\bf w}, b$\}, and $\omega$. Fortunately, given any two groups of parameters, the other group of parameters can be efficiently learned using either the stochastic gradient descent (SGD) algorithm (e.g. for \{${\bf w}, b$\} and $\omega$) or enumeration (e.g. for ${H}$). 

{ Therefore, we adopt a principled coordinate type algorithm to optimize the our deep structured model in Eq. (\ref{equ:svm_L2}) and (\ref{equ:svm_L3}). This learning algorithm actually is a general expectation maximization (GEM) method \citep{GEM1983}, which iteratively performs the E-step and the M-step: the former discovering the optimal latent variables by global searching and the latter optimizing the CNN and classifier parameters for a sub-optimal solution. As shown in \citep{GEM1983}, such a GEM procedure can converge monotonically to a stationary point.} More specifically, our learning algorithm iterates with the three steps: (i) Given the model parameters \{${\bf w}, b$\} and $\omega$, we estimate the latent variables $h_i$ for each video and update the corresponding feature $\phi(x_i; \omega, h_i)$ (Figure~\ref{fig:learning} (a)); { (ii) Given the updated features $\phi(X; \omega, H)$, we update the max-margin classifier \{${\bf w}, b$\} (Figure~\ref{fig:learning} (b)); (iii) Given the model parameters \{${\bf w}, b$\} and $H$, we update the CNN parameters $\omega$, which will lead to both the increase of the margin and the decrease of the radius (Figure~\ref{fig:learning} (c)).} {It is worth mentioning that the two steps (ii) and (iii) can be performed in the same procedure of SGD, i.e. their parameters are jointly updated in an end-to-end way.}

In the following, we explain in detail the three steps for minimizing the loss in Eq. (\ref{equ:svm_L3}), which are derived from our deep model.

{\textbf{(i)}} Given the model parameters $\omega$ and \{${\bf w}, b$\}, for each sample $(x_i, y_i)$, the most appropriate latent variables $h_i$ can be determined by exhaustive searching over all the possible choices,

\begin{eqnarray} \label{eq:EMH}
h_i^{*} = \arg\min_{h_i} {1 - \big({\bf w}\phi(x_i; \omega, h_i) + b\big)y_i}.
\end{eqnarray}
GPU programming is employed to accelerate the searching process. With the updated latent variables, we further obtain the feature set $\phi(X; \omega, H)$ of all the training data.

{\textbf{(ii)}} Given $\phi(X; \omega, H)$ and the CNNs' parameters $\omega$, batch stochastic gradient descent (SGD) is adopted for updating model parameters in Eq. (\ref{equ:svm_L3}). In iteration $t$, a batch ${B}_t \subset (X,Y,H)$ of $k$ samples is chosen. We can obtain the gradients of the max-margin classifier with respect to parameters \{${\bf w}, b$\},

\begin{small}
\begin{equation}
\label{equ:g_Wsvm}
\frac{\partial{L_3}}{\partial{{\bf w}}} = {\bf w} - \lambda \sum_{(x_i,y_i,h_i)\in {B}_t } y_i {\phi(x_i; \omega, h_i)} \max \bigg( 0, 1 - \big({\bf w}^T\phi(x_i; \omega, h_i) + b \big)y_i \bigg), \\
\end{equation}
\vspace{-10pt}
\begin{equation}
\label{equ:g_bsvm}
\frac{\partial{L_3}}{\partial b} = -2 \lambda \sum_{(x_i,y_i,h_i)\in {B}_t } y_i \max \bigg( 0, 1 - \big({\bf w}^T{\phi(x_i; \omega, h_i) + b}\big) y_i \bigg),
\end{equation}
\end{small}

{\textbf{(iii)}} Given the latent variables $H$ and the max-margin classifier \{${\bf w}, b$\}, based on the gradients with respect to $\omega$, the back propagation algorithm can be adopted to learn CNNs' parameters $\omega$. More specifically, we first update the mean $\bar{\phi}_{\omega}$ in Eq. (\ref{equ:def_mean}) based on $\phi(X; \omega, H)$, and then compute the derivative of the relaxed loss in Eq. (\ref{equ:svm_L3}) as

\begin{small}
\begin{equation}
\begin{split}
\label{equ:g_cnn_3}
&\frac{\partial{L_3}}{\partial{{\omega}}} = 4 \eta \sum_{(x_i,y_i,h_i)\in {B}_t } \left( \phi(x_i;\omega,h_i) - \bar{\phi}_{\omega} \right)^T  \frac{\partial \phi(x_i; \omega, h_i)}{\partial \omega} \\
& - 2 \lambda \sum{{\bf w}^Ty_i \frac{\partial{\phi(x_i;\omega, h_i)}}{\partial {{\omega}}}}\max \bigg( 0, 1 - \big({\bf w}^T\phi(x_i;\omega,h_i) + b \big)y_i \bigg).
\end{split}
\end{equation}
\end{small}

By performing the proposed  back propagation algorithm, we can further decrease the relaxed loss and optimize the model parameters. { During the back propagation, batch SGD is adopted to simultaneously update the parameters of both step (\textbf{ii}) and (\textbf{iii}).} The optimization algorithm iterates between these three steps until convergence.

\begin{small}
\begin{algorithm}[htb]
\caption{Learning Algorithm}
\label{alg:Framwork}
\begin{algorithmic}\footnotesize
\REQUIRE ~~\\
    The labeled 2D, 3D activity dataset and learning rate $\alpha_{{\bf w},b}$, $\alpha_{\omega}$.
\ENSURE ~~\\
    Model parameters \{$\omega, {\bf w}, b$\}.

\INPUT ~~\\
    Pre-train the spatio-temporal CNNs using the 2D videos.
    \\
\vspace{0.5em}
\hspace{-1.5em} Learning on 3D video dataset:
\MYWHILE
    \STATE
    \begin{itemize}
\setlength{\itemsep}{1pt}
 \setlength{\parskip}{0pt}
 \setlength{\parsep}{10pt}
      \item[1.] Estimate the latent variables ${H}$ for all samples by fixing model parameters \{$\omega, {\bf w}, b$\}.
      \item[2.] Optimize \{${\bf w}, b$\} given the CNN model parameters $\omega$ and the input sample segments indicated by ${H}$:
      \begin{itemize}
      \item[2.1] Calculate $\phi(X;\omega,H)$ by forwarding the neural network with $\omega$.
      \item[2.2] Optimize \{${\bf w}, b$\} via:

      ${\bf w}$ := ${\bf w} - \alpha_{{\bf w}, b} * \frac{\partial{L}}{\partial {\bf w}}$ by Eq.~(\ref{equ:g_Wsvm});

	  $b$ := $b - \alpha_{\bf w, b} * \frac{\partial{L}}{\partial b}$ by Eq.~(\ref{equ:g_bsvm});

      \end{itemize}

      \item[3.] Optimize $\omega$ given \{${\bf w}, b$\} and ${H}$:
      \begin{itemize}
      \item[3.1] Calculate $\kappa_{ij}$, $\kappa_i$ and $\phi_i$ for $L_2$, or calculate $\bar{\phi}_{\omega}$ for $L_3$.
	  \item[3.2] Optimize the parameters $\omega$ of the spatio-temporal CNNs:

	  $\omega$ := $\omega - \alpha_{\omega} * \frac{\partial{L}}{\partial{\omega}}$ by Eq.~(\ref{equ:g_cnn_3}).
      \end{itemize}

    \end{itemize}
\MYENDWHILE {$L$ in (\ref{equ:svm_L2}) or (\ref{equ:svm_L3}) converges. }

\end{algorithmic}
\end{algorithm}
\end{small}

\subsection{Model Pre-training}

Parameter pre-training followed by fine-tuning is an effective method to  boost the performance in deep learning, especially when the training data is scarce. In the literature, there are two popular solutions, i.e. unsupervised pre-training on unlabeled data~\citep{sermanet-cvpr-2013} and supervised pre-training for an auxiliary task~\citep{rcnn2014}. The latter usually requires the data formate (e.g. image) for parameter pre-training is exactly the same as that (e.g. image) for fine-tuning.

In our approach, we suggest an alternative solution for 3D human activity recognition. Although collecting RGB-D videos of human activities is expensive,  a large amount of 2D activity videos can be easily obtained. Consequently, we first apply the supervised pre-training using a large number of 2D activity videos, and then fine-tune the CNNs' parameters for training the 3D human activity models.

In the step of pre-training, the CNNs' parameters are randomly initialized at the beginning. For each input 2D video, we equally segment it into $M$ parts without estimating its latent variables. { Here we simply employ the softmax classifier to pre-train the parameters of CNN, since the softmax loss unbiasedly treat all samples and it is suitable for learning a general feature representation \citep{rcnn2014}.}

The 3D and 2D convolutional kernels obtained in pre-training are only for gray channel. Thus, after pre-training, we duplicate the dimension of the 3D convolutional kernels in the first layer and initialize the parameters for the depth channel by the parameters for the gray channel, which allows us to borrow the features learned from the 2D video while directly learning the higher level information from the specific 3D activity dataset. For the fully connected layer, we set its parameters as random values.

We summarize the overall learning procedure in Algorithm \ref{alg:Framwork}.

%\begin{figure*}[!htb]
%\centering
%%\epsfig{figure=./images/dataset,width=\textwidth}
%\includegraphics[width= 1.6\columnwidth]{./images/sbu}
%\caption{Activity examples from the {\em SBU} databases. From left to right and top to bottom, these categories are \{{\em  departing, pushing, approaching, exchanging objects, kicking, hugging, punching shaking hands}\}, respectively.}\label{fig:sbu}
%\end{figure*}

\section{Inference}
Given an input video $x_i$, the inference task aims to recognize its category of the activity, which can be formulated as the minimization of $F_y(x_i,\omega,h)$ with respect to the activity label $y$ and the latent variables $h$,

\begin{eqnarray} \label{eq:inference}
(y^{*},h^{*}) = \arg \max_{(y,h)} \{ F_y(x_i,\omega,h) = {\bf w}_y^T \phi(x_i; \omega, h) + b_y \}.
\end{eqnarray}
where \{${\bf w}_y, b_y$\} denotes the parameters of the max-margin classifier for the activity category $y$. Note the possible values for $y$ and $h$ are discrete. Thus the problem above can be solved by searching across all of the labels $y (1 \leq y \leq C)$ and calculate the maximum $F_y(x_i,\omega,h)$ by optimizing $h$. To find the maximum of $F_y(x_i,\omega,h)$, we enumerate all the possible values of $h$, and calculate the corresponding $F_y(x_i,\omega,h)$ via forward propagations. Since the forward propagations decided by different $h$ are independent, we can parallelize the computation via GPU to accelerate the inference process.

\begin{figure*}[!htb]
\centering
\includegraphics[width= \textwidth]{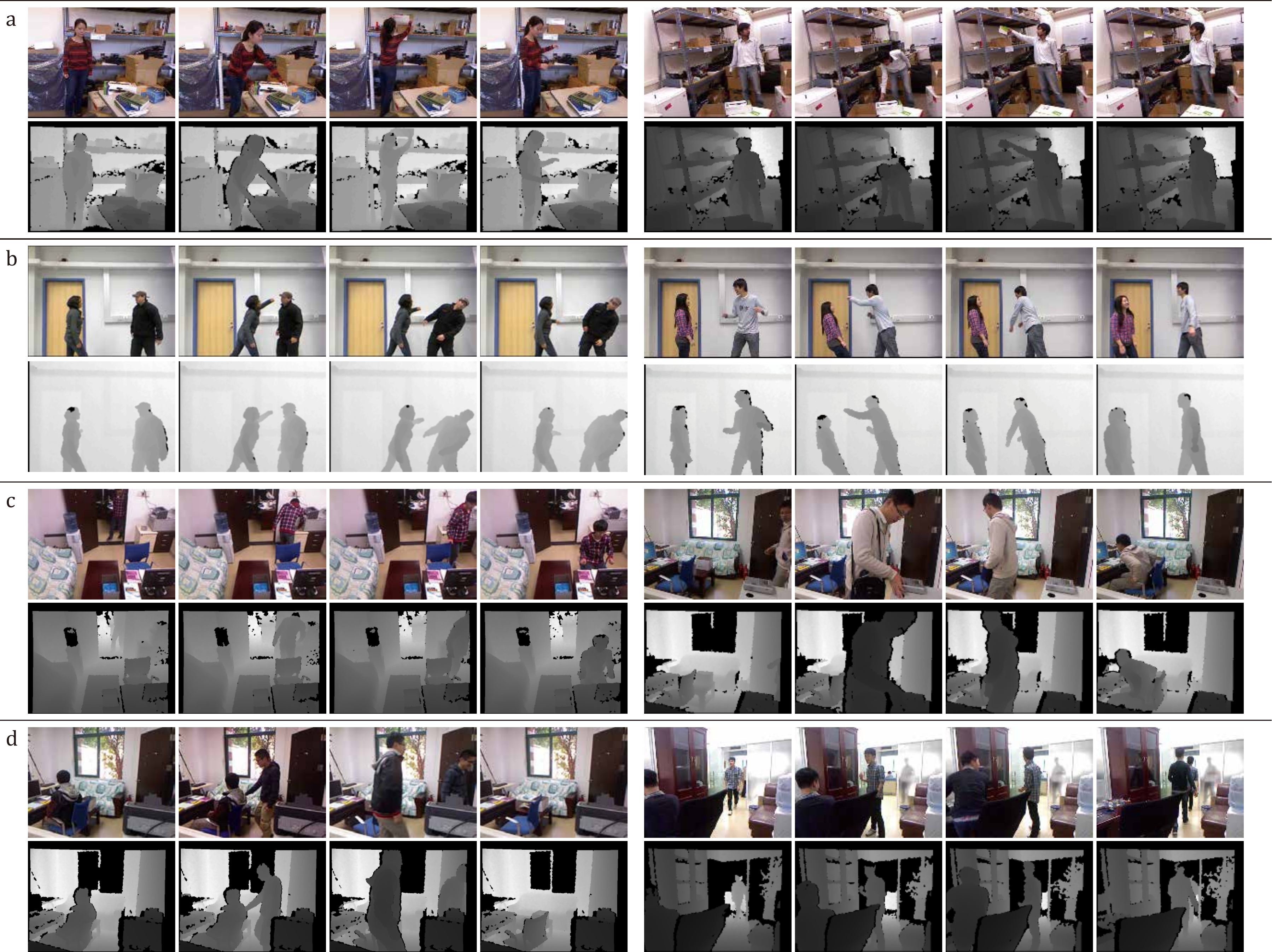}
\caption{{Activity examples from the testing databases. Several sampled frames and depth maps are presented. (a) {\em CAD-120}, (b) {\em SBU}, (c) {\em OA1}, (d) {\em OA2}, respectively, show two activities of the same category selected from the three databases. }}\label{fig:cad120}\label{fig:sbu}\label{fig:OA1}\label{fig:OA2}
\end{figure*}

\section{Experiments}
\label{sec:exper}

To validate the advantages of our model, experiments are conducted on several challenging public datasets, i.e. {\em CAD-120 Dataset}~\citep{CADIJRR2013}, {\em SBU Kinect Interaction Dataset}~\citep{SBU}, and a larger dataset newly created by us, namely {\em Office Activity (OA) Dataset}. { Moreover, we introduce a more comprehensive dataset in our experiments by combining five existing datasets of RGB-D human activity. In addition to demonstrating the superior performance of the proposed model over other state-of-the-arts, we extensively evaluate the main components of our framework.}

\subsection{Datasets and Setting}

The {\em CAD-120} dataset comprises of 120 RGB-D video sequences of humans performing long daily activities of $10$ categories, and has been widely used for testing 3D human activity recognition methods. These activities recorded via the Microsoft Kinect sensor were performed by four different subjects, and each activity was repeated three times by the same actor. These activities have a long sequence of sub-activities, which vary from subject to subject significantly in terms of length of the sub-activities, order of the sub-activities as well as in the way they executed the task. Moreover, the challenges on this dataset also lie in the large variance in object appearance, human pose, and viewpoint. Several sampled frames and depth maps from this databases of these $10$ categories are exhibited in Figure~\ref{fig:cad120} (a).

The {\em SBU} dataset consists of $8$ categories of two-person interaction activities, including a total of about 300 RGB-D video sequences, i.e. about $40$ sequences for each interaction category. Even though most interactions in this dataset are simple, it is still challenging for modeling two-person interactions by considering the following difficulties: i) one person is acting and the other person is reacting in most cases, ii) the average frame length of these interaction is short (ranging from 20 to 40), iii) the depth maps have noises. Figure~\ref{fig:sbu} (b) shows several sampled frames and depth maps of these $8$ categories.

%\begin{figure*}[!htb]
%\centering
%\psfig{figure=./images/dataset,width=\textwidth}
%\includegraphics[width= 1.6 \columnwidth]{./images/OA1}
%\caption{Activity examples from the {\em OA1} database. From left to right and top to bottom, these categories are \{{\em answering-phones, arranging-files, eating, moving-objects, going-to-work, finding-objects, mopping, sleeping, taking-water, wandering}\}, respectively.}\label{fig:OA1}
%\end{figure*}

The proposed {\em OA} dataset is more comprehensive and challenging compared with the existing datasets, and it covers the regular daily activities taken place in an office. To the best of our knowledge, it is the largest activity dataset of RGB-D videos consisting of $1180$ sequences. The {\em OA} database is publicly accessible \footnote{\url{http://vision.sysu.edu.cn/projects/3d-activity/}}. Three RGB-D sensors (i.e. Microsoft Kinect cameras) are utilized to capture data from different viewpoints, and more than 10 actors are involved. The activities are captured in two different offices to increase the variability, where each actor performs the same activity twice. The activities performed by two subjects with interactions are also included. Specifically, it is divided into two subsets, each of which contains 10 categories of activities: {\em OA1} (complex activities by a single subject) and {\em OA2} (complex interactions by two subjects). Several sampled frames and depth maps are exhibited in Figure~\ref{fig:OA1} (c) and Figure~\ref{fig:OA2} (d) from {\em OA1} and {\em OA2}, respectively.

{
To evaluate our model under a larger scale scenario, we collect an extra dataset by combining existing RGB-D human activity datasets: { {\em RGBD-HuDaAct}~\cite{ni2013rgbd},  {\em CAD120}, {\em SBU}, {\em UTKinect-Action}~\cite{HOJ3D} and {\em OA}}. This dataset contains $2989$ video sequences with 5, 500, 000 frames (approximately 50 hours long) belonging to $50$ activity categories, and we name it as {\em Merged\_50} Dataset. Note that we merge very similar activity categories from the different datasets. In addition, we create a coarse-level variant of this dataset by merging the $50$ categories into only $4$, that is, all of the $2989$ activity instances are roughly divided into $4$ types: \{a person interacting small objects (e.g. answering-phones, having-meal), a person interacting large objects (e.g. sleeping-in-bed, cleaning-objects), physical contacting of persons (e.g. departing, asking-and-way), non - physical contacting of persons (e.g. exchanging objects, hugging objects together)\}. And we name this coarse-level dataset as {\em Merged\_4}.
}

\begin{table*}[!htb]
\center
\begin{tabular}{|c|c|c|c|c|c|c|c|}
\hline
\hline
& ~\citep{3DCNNPAMI} & Softmax & SVM  & R-SVM & Softmax & SVM & R-SVM \\
& & + CNN & + CNN & + CNN & + LCNN & + LCNN & + LCNN \\
\hline
without pre-training & 56.4\% & 61.8\% & 57.5\% & 62.5\% & 70.8\% & 66.7\% & \textbf{74.7}\% \\
\hline
with pre-training & 63.1\% & 68.3\% & 78.3\% & 77.7\% & 82.7\% & 89.4\% & \textbf{90.1}\% \\
\hline
\hline
\end{tabular}
\caption{{ Average accuracy with/without incorporating the latent structure on {\em CAD 120} dataset with different top classifiers: Softmax, linear SVM, radius-margin SVM}.}
\vspace{-10pt}
\label{tab:latentStruct}\label{tab:pretrain}
\end{table*}

\begin{figure}[!htb]
\centering
\includegraphics[width=\columnwidth]{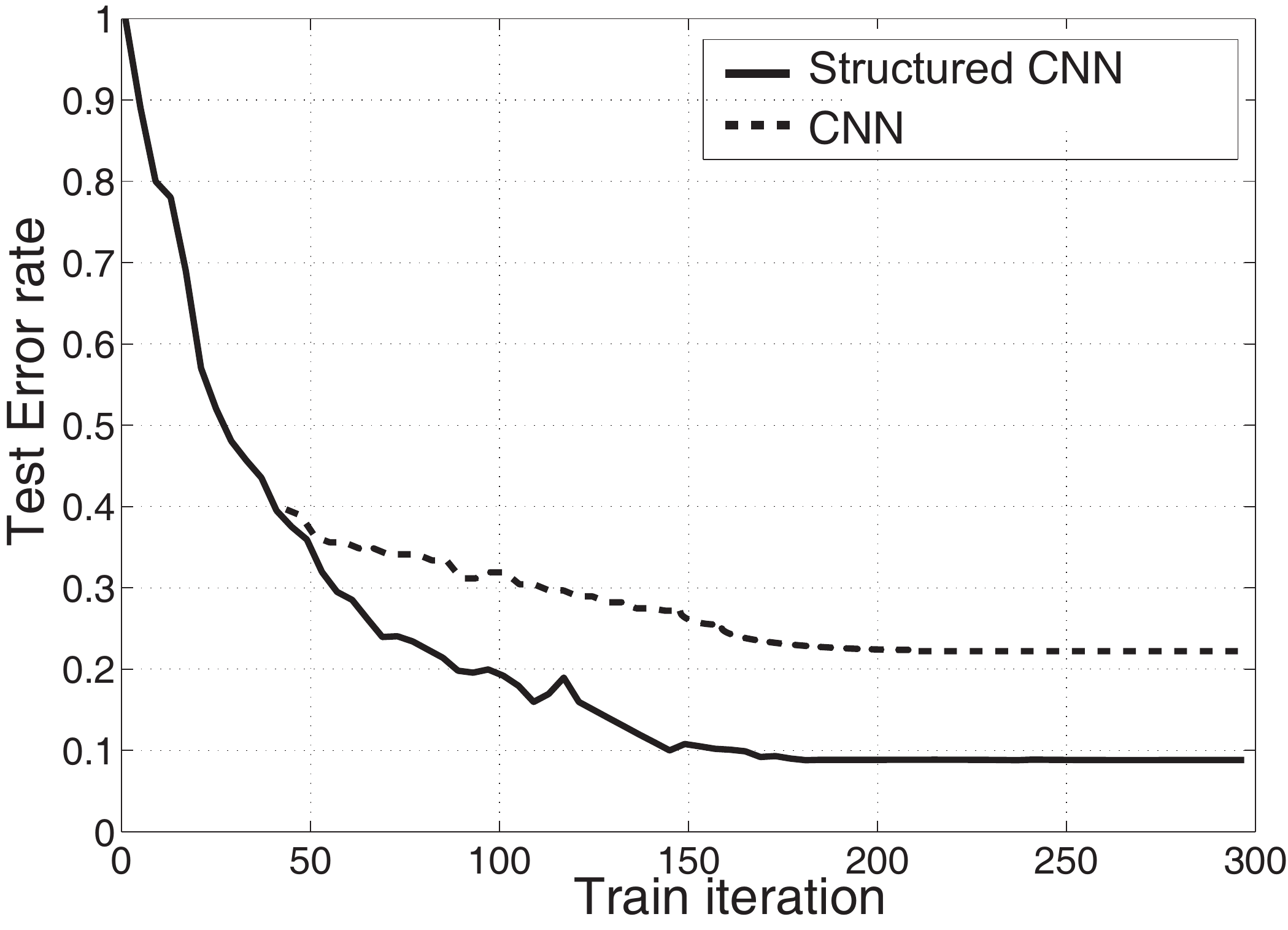}
\caption{{ Test error rates with/without incorporating the latent structure in the deep model. The solid curve represents the deep model trained by the proposed joint component learning method, and the dashed curve represents the traditional training way (i.e. using standard back-propagation).} }\label{fig:latentStruct}
\end{figure}

All the experiments are executed on a desktop PC with an {Intel i7 4.0GHz CPU, 8GB RAM and GTX 980 GPU}. For model learning, Algorithm~\ref{alg:Framwork} is employed to learn the CNN $\omega$ and the classifier \{${\bf w}, b$\}. { The main time-consuming part is the model pre-training, which can take several days based on our desktop PC. Afterwards, the training time of our model on 3D activity datasets is acceptable: 3 hours on CAD-120 (including 120 long videos). Each iteration of training costs similar time, and the convergence of our model over iterations is shown in Figure~\ref{fig:latentStruct}.} For inference, with the GPU-based parallel implementation, it only takes around $0.4$ seconds to complete recognition on a given video with about 200 frames.
For {\em CAD-120} and {\em SBU}, we follow the same training/test split adopted in the comparison methods. For {\em OA1} and {\em OA2}, we adopt the 5-fold cross validation by ensuring that the subjects in training set are different with those in testing set. { Since {\em Merged\_50} and {\em Merged\_4} datasets contain different subjects in different environments, we randomly select 70\%, 10\%, 20\% video sequences from the two datasets for training, validating and testing, respectively.}

\subsection{Empirical Analysis}

Empirical analysis are given to assess the main components of the proposed deep structured model, including latent structure, relaxed radius-margin bound, model pre-training, and depth/grayscale channel. Several variants of our method are suggested by enabling/disabling some components. Specifically, we denote the conventional 3D {convolutional} neural network with the softmax classifier as Softmax + CNN, denote the 3D CNN with the SVM classifier as SVM + CNN, denote the 3D CNN with the relaxed radius-margin bound classifier as R-SVM + CNN. Analogously, we denote our deep model as LCNN, and then define Softmax + LCNN, SVM + LCNN, and R-SVM + LCNN accordingly.

{
{\bf Latent Model Structure.} In this experiment, we implement a simplified version of our model by removing the latent structure and compare it with our full model. The simplified model is actually a spatio-temporal CNN model including both 3D and 2D convolutional layers, and this model uniformly segments the input video into $M$ sub-activities. Without the latent variables to be estimated, the standard back propagation algorithm is employed for model training. We execute this experiment on {\em CAD120} dataset. Figure~\ref{fig:latentStruct} shows the test error rates with different iterations of 
{ the simplified model (i.e. CNN) and the full version (i.e.  structured CNN) in the same CNN initialization}. Based on the results, we observe that our full model outperforms the simplified model in both error rate and training efficiency. Furthermore, one can see that the structured models with model pre-training, i.e. Softmax + LCNN, SVM + LCNN, R-SVM + LCNN, achieve {$14.4\%/11.1\%/12.4\%$} better performance than the traditional CNN models, i.e. Softmax + CNN, SVM + CNN, R-SVM + CNN, respectively. The results clearly demonstrate the significance of incorporating latent temporal structure in dealing with the large temporal variations of human activities.}

%\begin{figure}[!htb]
%\centering
%%\psfig{figure=pretrain,width=\columnwidth}
%\includegraphics[width=\columnwidth]{pretrain}
%\caption{{ Test error rates with/without using the pre-training}.}\label{fig:pretrain}
%\end{figure}

{\bf Pre-training.} To justify the effectiveness of pre-training, we discard the parameters trained on 2D videos, and learn the model directly on the grayscale-depth data. %Then we compare the models with/without pre-training by the test error rate. 
%To analyze the rate of convergence, we adopt the R-SVM + LCNN framework, and let with/without pre-training share the same learning rate settings for fair comparison.  
%Using the {\em CAD120} dataset, we plot the test error rates with the increasing of iteration numbers during training in Figure~\ref{fig:pretrain}. It is shown that the model using pre-training converges in $170$ iterations while the one without pre-training requires $300$ iterations, and the model with pre-training converges to a much lower test error rate (9\%) that that (25\%) without pre-training. Furthermore, 
We compare the performance with/without pre-training using SVM + LCNN and R-SVM + LCNN, as listed in Table~\ref{tab:pretrain}. One can see that pre-training is effective in reducing the test error rate. Actually, the test error rate with pre-training is about 15\% less than that without pre-training.

\begin{table*}
\center
\begin{tabular}{|c|c|c|c|c|c|c|c|}
\hline
\hline
& \citep{SungICRA2012} & best result from \citep{CADIJRR2013} & \citep{DSTIP} & \citep{3DCNNPAMI} & best result from \citep{SaxenaICML2013}  & R-SVM + LCNN \\
\hline
arranging-objects & - & 33.0\% & 75.0\% & 68.3\% & 50.0\% & \textbf{91.7}\% \\
cleaning-objects  & - & 67.0\% & 68.3\% & 60.0\% & 67.0\% & \textbf{83.3}\%\\
having-meal  & - & \textbf{100.0}\% & 41.7\%  & 60.0\% & \textbf{100.0}\% & 91.7\%\\
making-cereal  & - & \textbf{100.0}\%  & 76.7\% & 77.6\% & \textbf{100.0}\% & \textbf{100.0}\%\\
microwaving-food  & - & \textbf{100.0}\% & 36.7\% & 71.7\% & 67.0\% & \textbf{100.0}\% \\
picking-objects  & - & 75.0\% & 75.0\% & 58.3\% & 67.0\% & \textbf{91.7}\%\\
stacking-objects  & - & \textbf{92.0}\% & 75.0\% & 48.3\% & \textbf{92.0}\% & 91.7\%\\
taking-food  & - & 75.0\% & 83.3\% & 73.3\% & 67.0\% & \textbf{91.7}\%\\
taking-medicine  & - & \textbf{100.0}\% & 58.3\%  & 76.7\% & {92.0}\% & 84.6\%\\
unstacking-objects  & - & \textbf{100.0}\% & 33.3\% & 36.7\% & {92.0}\% & 75.0\%\\
\hline
\hline
Average accuracy & 59.7\% & 84.2\% & 62.3\% & 63.1\% & 83.1\% & \textbf{90.1}\%\\
\hline
\hline
\end{tabular}
\caption{{ Accuracy of all categories on {\em CAD120} dataset. Accuracy per activity category and average accuracy of all categories are reported.}}
\label{tab:cad120_all} \label{tab:cad120_avg}
\end{table*}

\begin{figure*}[!htb]
\centering
\includegraphics[width=\textwidth]{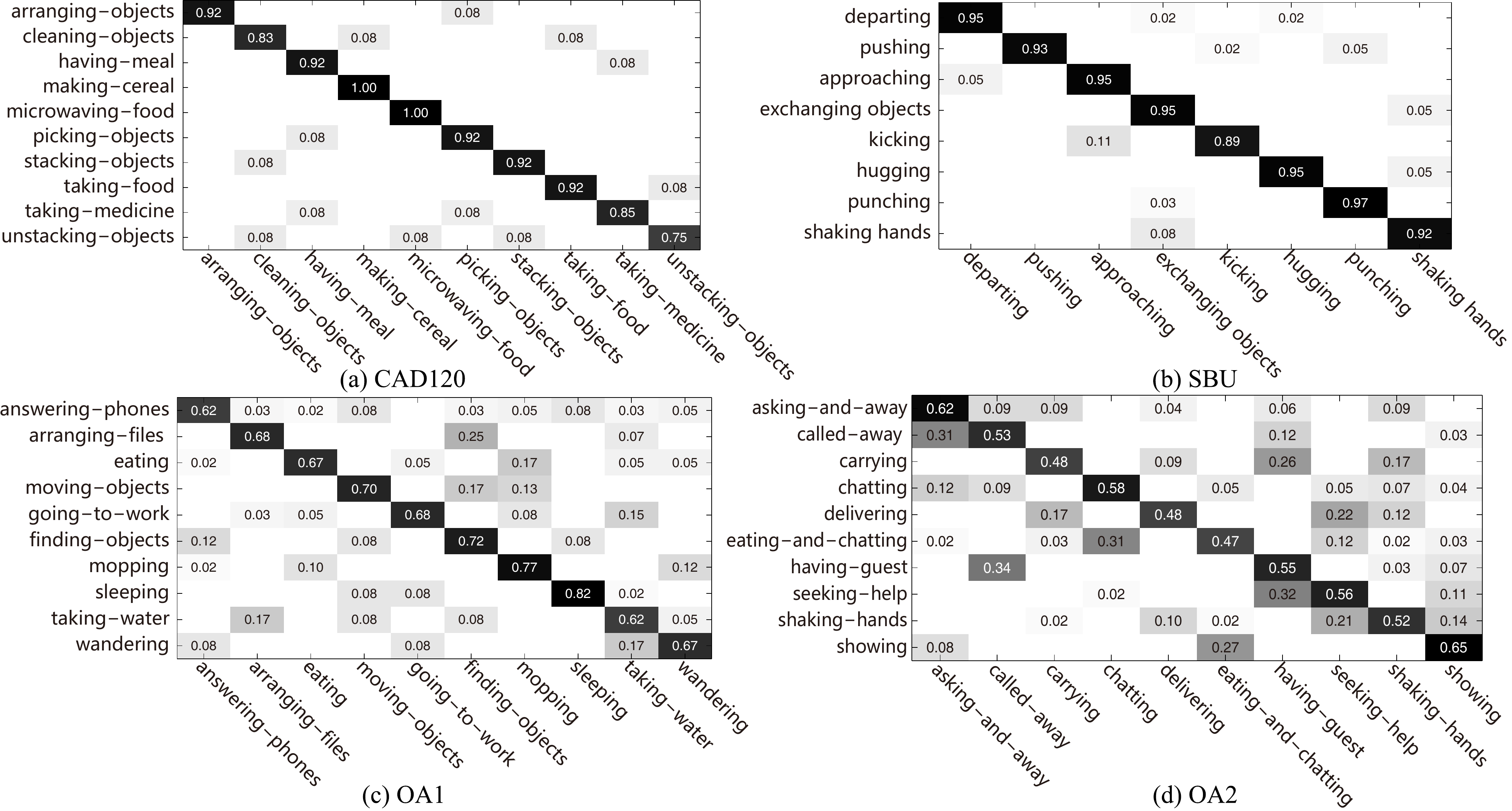}
\caption{{ Confusion Matrices of our proposed deep structured model on (a) {\em CAD120}, (b) {\em SBU}, (c) {\em OA1}, (d) {\em OA2} datasets. It is evident that these confusion matrices all have a strong diagonal with few errors}.}\label{fig:confusionMat}
\vspace{-10pt}
\end{figure*}

{\bf Relaxed Radius-margin Bound.} As described above, the training data for grayscale-depth human activity recognition are scarce. Thus, for the last fully connected layer, we adopt the SVM classifier by incorporating with the relaxed radius-margin bound, resulting in the R-SVM + LCNN model. { To justify the role of the relaxed radius-margin bound, Table~\ref{tab:r-svm} compares the accuracies of Softmax + LCNN, SVM + LCNN, and R-SVM + LCNN on all datasets with the same experimental settings. It is observed that the max-margin based classifiers (SVM and R-SVM) are particularly effective on small scale datasets (CAD120, SBU, OA1, OA2, Merged\_50). On average, the accuracy of R-SVM + LCNN is average $6.5\%$ higher than that of Softmax + LCNN, and is about $1\%$ higher than that of SVM + LCNN. On Merged\_4 dataset, the improvement of R-SVM + LCNN is incrementally evident, $1.8\%$ higher than Softmax + LCNN. These results finely accord with our motivation of incorporating the radius-margin bound into our deep learning framework.} Moreover, when the model is learned without pre-training, R-SVM + LCNN gains about $4\%$ and $8\%$ improvements over Softmax + LCNN and SVM + LCNN by accuracy, respectively, as Table~\ref{tab:pretrain} reports.

\begin{table}
\center
\begin{tabular}{|c|c|c|c|}
\hline
\hline
& Softmax + & SVM + & R-SVM + \\
& LCNN & LCNN & LCNN \\
\hline
CAD120 & 82.7\% & 89.4\% & \textbf{90.1}\% \\
SBU & 92.4\% & 92.8\% & \textbf{94.0}\% \\
OA1 & 60.7\% & 68.5\% & \textbf{69.3}\% \\
OA2 & 47.0\% & 53.7\% & \textbf{54.5}\% \\
Merged\_50 & 30.3\% & 36.4\% & \textbf{37.3}\% \\
Merged\_4 & 87.1\% & 88.5\% & \textbf{88.9}\% \\
\hline
\hline
\end{tabular}
\caption{{ Average accuracy of all categories on four datasets with different classifiers.}}
\label{tab:r-svm}
\end{table}

\begin{table}[!htbp]
\center
\begin{tabular}{|c|c|c|c|}
\hline
\hline
  & grayscale & depth & grayscale + depth \\
\hline
OA1 & 60.4\% & 65.2\% & \textbf{69.3}\% \\
OA2 & 46.3\% & 51.1\% & \textbf{54.5}\% \\
Merged\_50 & 27.8\% & 33.4\% & \textbf{37.3}\% \\
Merged\_4 & 81.7\% & 85.5\% & \textbf{88.9}\% \\
\hline
\hline
\end{tabular}
\caption{{ Channel analysis on the three datasets. Average accuracy of all categories are reported.}}
\label{tab:channel}
\end{table}

{\bf Channel Analysis.} To evaluate the contribution of the grayscale and depth data, we execute the following experiment on the {\em OA} datasets: keeping only one channel data as input. Specifically, we first disable the depth channel and input the grayscale data to perform the training/testing, and then disable the grayscale channel and employ the depth channel for training/testing. { Table~\ref{tab:channel} proves that depth data can boost the performance by large margins, especially in {\em OA1} and {\em Merged\_50}}. This is due to the fact that large appearance variances existed in grayscale data. In particular, our testing is performed on the new subjects and this would further increase the appearance variance. On the contrary, the depth data are more reliable and have much smaller variances, which is helpful in capturing the salient motion information.

\begin{table}
\center
\begin{tabular}{|c|c|c|c|}
\hline
\hline
& Linear SVM & MILBoost & ours \\
\hline
Average accuracy & 87.3\% & 91.1\% & \textbf{93.4}\%\\
\hline
\hline
\end{tabular}
\caption{{ Average accuracy on {\em SBU} dataset.}}
\label{tab:sbu}
\end{table}

\begin{table}[!hbp]
\center
\begin{tabular}{|c|c|c|c|}
\hline
\hline
&\citep{DSTIP} &\citep{3DCNNPAMI} & ours \\
\hline
answering-phones & 12.5\% & 40.0\% & \textbf{61.7}\% \\
arranging-files & 59.7\% & 53.3\% & \textbf{68.3}\% \\
eating & 40.3\% & 41.7\% & \textbf{66.7}\%\\
moving-objects & 48.6\% & 51.7\% & \textbf{70.0}\%\\
going-to-work & 34.7\% & 41.7\% & \textbf{68.3}\%\\
finding-objects & 65.3\% & 36.7\% & \textbf{71.7}\% \\
mopping & 63.9\% & 66.7\% & \textbf{76.7}\%\\
sleeping & 25.0\% & 45\% & \textbf{81.7}\%\\
taking-water & 58.3\% & 40.0\% & \textbf{61.7}\%\\
wandering & 56.9\% & 50.0\% & \textbf{66.7}\%\\
\hline
\hline
Accuracy & 46.5\% & 46.7\% & \textbf{69.3}\%\\
\hline
\hline
\end{tabular}
\caption{{ Quantitative results on {\em OA1} dataset.  Accuracy per activity category and average accuracy of all categories are reported.}}
\label{tab:oa1}
\end{table}

\begin{table}[!hbp]
\center
\begin{tabular}{|c|c|c|c|}
\hline
\hline
&\citep{DSTIP} &\citep{3DCNNPAMI} & ours \\
\hline
asking-and-away & 12.5\% & 39.6\% & \textbf{62.3}\% \\
called-away & 45.8\% & 44.8\% & \textbf{53.5}\%\\
carrying & \textbf{66.7}\% & 56.8\% & 48.3\%\\
chatting & 37.5\% & 17.2\% & \textbf{57.9}\%\\
delivering & 20.1\% & 34.5\% & \textbf{48.3}\%\\
eating-and-chatting & \textbf{50.0}\% & 35.8\% & 46.6\% \\
having-guest & 37.5\% & 34.1\% & \textbf{55.2}\%\\
seeking-help & 16.7\% & 44.8\% & \textbf{56.1}\% \\
shaking-hands & 41.7\% & 32.8\% & \textbf{51.7}\%\\
showing & 37.5\% & 29.3\%  & \textbf{64.6}\%\\
\hline
\hline
Accuracy & 36.6\% & 37.0\% & \textbf{54.5}\%\\
\hline
\hline
\end{tabular}
\caption{ { Quantitative results on {\em OA2} dataset. Accuracy per activity category and average accuracy of all categories are reported}.}
\label{tab:oa2}
\end{table}

\begin{table}
\center
\begin{tabular}{|c|c|c|c|}
\hline
\hline
&\citep{DSTIP} &\citep{3DCNNPAMI} & ours \\
\hline
\hline
Merged\_50 & 21.1\% & 24.1\% & \textbf{37.3}\%\\
Merged\_4 & 79.1\% & 81.2\% & \textbf{88.9}\%\\
\hline
\hline
\end{tabular}
\caption{{ Average accuracy on {\em Merged\_50} and {\em Merged\_4} datasets.}}
\label{tab:merged}
\end{table}

%\begin{figure*}[!htb]
%\centering
%%\epsfig{figure=./images/dataset,width=\textwidth}
%\includegraphics[width= 1.6 \columnwidth]{./images/OA2}
%\caption{Activity examples from the {\em OA2} database. From left to right and top to bottom, these categories are \{{\em asking-and-away, called-away, carrying, chatting, delivering, eating-and-chatting, having-guest, seeking-help, shaking-hands, showing}\}, respectively.}\label{fig:OA2}
%\end{figure*}

\subsection{Experimental Results and Comparisons}

{\em CAD-120 dataset.} On this dataset, we adopt five state-of-the-art methods for comparison. Note that for different methods we train the models using the same data annotation, which only includes the activity labels on videos. As shown in Table~\ref{tab:cad120_avg}, our method obtains the average accuracy of $90.1\%$, which is significantly superior to the results generated by other five competing methods, i.e. $59.7\%$~\citep{SungICRA2012}, $84.2\%$~\citep{CADIJRR2013}, $62.3\%$~\citep{DSTIP}, $63.1\%$~\citep{3DCNNPAMI} and $83.1\%$~\citep{SaxenaICML2013}. Table~\ref{tab:cad120_all} reports the accuracies per activity category of our method and the method based on hand-crafted feature engineering~\citep{DSTIP}, the deep architecture of convolutional neutral networks~\citep{3DCNNPAMI} \footnote{ { We implement the 3D-CNN model~\cite{3DCNNPAMI}. For fair comparison, parameter pre-training and dropout have been also employed in our implementation, and the configuration of 3D-CNN is the same with that of our model except that we set $M = 1$ for 3D-CNN.}}, and the rich spatio-temporal relations modeling~\citep{SaxenaICML2013}. Our method achieves the highest accuracies on $6$ of the 10 activity categories.

{\em SBU dataset.} As shown in Table~\ref{tab:sbu}, our method obtains the average accuracy of $93.4\%$ and performs better than the methods based on body-pose features~\cite{SBU}, which indicates that our method is effective in learning discriminative features directly from raw data for modeling person-to-person interaction.

{\em OA dataset.} In this experiment, we apply our method on the two {\em OA} subsets. Tables.~\ref{tab:oa1} and~\ref{tab:oa2} list the accuracies per category and average accuracy of the competing methods, and our method outperforms the state-of-the-art methods in terms of the average accuracy. On the {\em OA1} set, our method achieves the best accuracies on all  categories and obtains the highest average accuracy of $69.3\%$, as shown in Table.~\ref{tab:oa1}. On the {\em OA2} set, our method achieves the best accuracies on 8 out of 10 activity categories and obtains the highest average accuracy of $54.5\%$, as shown in Table~\ref{tab:oa2}. By checking the results, we find that the failure cases are mainly caused by the lack of contextualized scene understanding. For example, understanding the activities of {\em having-guest} and {\em eating-and-chatting} actually requires extra higher level information, and we will consider this issue in the future work.

{
{\em Merged datasets.} Table~\ref{tab:merged} reports the average accuracy of the competing methods, and our method outperforms the state-of-the-art methods~\citep{DSTIP, 3DCNNPAMI} in terms of the average accuracy.
}

In summary, our method consistently achieves better results than the competing methods on the three 3D activity datasets. Figure~\ref{fig:confusionMat} shows the confusion matrices of our model for all datasets. One can see that, the confusion matrices are strongly diagonal with few errors, which indicates that our deep structured model is effective in handling various challenges in 3D human activity recognition.

\section{Conclusion}

In this paper, we have introduced, first, a deep
and latent-structured model using the convolutional neural networks.
Second, a unified formulation integrating the radius-margin regularization with the feature learning. Third, an effective learning algorithm that iteratively optimizes the sub-activity decomposition, the margin-based classifier, and the neural networks. We have demonstrated the practical applicability of our model by effectively recognizing human activities using a depth camera. Experiments on the public datasets suggest that our model convincingly outperforms other state-of-the-art methods under several very challenging scenarios.

{ One main drawback of our current solution is the scalability of model inference. The brute-force enumeration over all settings of the latent variables will cause extra computation cost and this issue may become much more serious when the number (e.g., 1000) of human activity categories is large. Apart from the scalability issue}, we intend to extend our work in the following directions. The first is to generalize our model with compositional grammar rules (e.g. the And-Or grammars), and thus deal with more complicated event understanding (e.g. the causality inference). { The second is to revise our neural network for recognizing human action / activity from 2D videos. Note that there are distinct differences between 2D videos and 3D videos. For example, these mentioned 2D datasets basically include diverse environments (e.g., indoor / outdoor) with the camera moving, and the 3D depth data are all captured indoor with a fixed sensor (i.e. Microsoft Kinects). In addition, the 2D videos are usually in higher resolution than the data (i.e. 320 $\times$ 240) captured by the depth sensor.}

\section*{Acknowledgment}
This work was supported in part by the Hong Kong Scholar Program, and in part by the HK PolyU's Joint Supervision Scheme with the Chinese Mainland, Taiwan and Macao Universities (Grant no. G-SB20), in part by Guangdong Natural Science Foundation (Grant no. s2013010013432), and in part by Guangdong Science and Technology Program (Grant no. 2013B010406005).

\bibliographystyle{spbasic}

\end{document}